\documentclass[15pt]{article}

\usepackage{arxiv}
\usepackage[utf8]{inputenc} 
\usepackage[hidelinks]{hyperref}
\usepackage{booktabs}       
\usepackage{amsfonts}       
\usepackage{nicefrac}       
\usepackage{microtype}      
\usepackage{graphicx}
\usepackage[numbers]{natbib}
\usepackage{doi}
\usepackage{relsize}
\usepackage{amssymb}
\usepackage{amsmath}
\usepackage{lineno}
\usepackage{multirow}
\usepackage{bigints}
\usepackage{amsbsy}
\usepackage{dcolumn}
\usepackage{bm}
\usepackage{cleveref}       
\usepackage[utf8]{inputenc}
\usepackage[T1]{fontenc}
\usepackage{mathptmx}
\usepackage{etoolbox}
\usepackage{xcolor}
\usepackage{subfig}
\usepackage{calrsfs}
\usepackage{bm}
\DeclareMathAlphabet{\pazocal}{OMS}{zplm}{m}{n}
\def \PCE {\textrm{\scalebox{1.05}{PCE}}}
\def \PC {\textrm{\scalebox{1.05}{PC}$^2$}}
\def \PIML {\textrm{\scalebox{1.05}{SciML}}}
\def \UQ {\textrm{\scalebox{1.05}{UQ}}}

\usepackage[utf8]{inputenc}

\title{Physics-constrained polynomial chaos expansion for scientific machine learning and uncertainty quantification}

\date{}

\newif\ifuniqueAffiliation
\uniqueAffiliationtrue

\ifuniqueAffiliation 
\author{
        Himanshu Sharma\\
	Department of Civil and Systems Engineering\\
	  Johns Hopkins University\\
	Baltimore, USA  \\
	\And
	Luk{\'a}{\v s} Nov{\'a}k\\
	Department of Civil Engineering\\
	Brno University of Technology\\
        Brno, Czech Republic\\
        \And
        Michael Shields \thanks{Corresponding author;\  email: \texttt{michael.shields@jhu.edu}} \\
	Department of Civil and Systems Engineering\\
	  Johns Hopkins University\\
	Baltimore, USA  \\
 }

\begin{document}
\maketitle
\begin{abstract}
We present a novel physics-constrained polynomial chaos expansion as a surrogate modeling method capable of performing both scientific machine learning (\PIML{}) and uncertainty quantification (\UQ{}) tasks. The proposed method possesses a unique capability: it seamlessly integrates \PIML{} into \UQ{} and vice versa, which allows it to quantify the uncertainties in \PIML{} tasks effectively and leverage \PIML{} for improved uncertainty assessment during \UQ{}-related tasks. 
The proposed surrogate model can effectively incorporate a variety of physical constraints, such as governing partial differential equations (PDEs) with associated initial and boundary conditions constraints, inequality-type constraints (e.g., monotonicity, convexity, non-negativity, among others), and additional a priori information in the training process to supplement limited data. This ensures physically realistic predictions and significantly reduces the need for expensive computational model evaluations to train the surrogate model. Furthermore, the proposed method has a built-in uncertainty quantification (UQ) feature to efficiently estimate output uncertainties. To demonstrate the effectiveness of the proposed method,  we apply it to a diverse set of problems, including linear/non-linear PDEs with deterministic and stochastic parameters, data-driven surrogate modeling of a complex physical system, and \UQ{} of a stochastic system with parameters modeled as random fields.

\end{abstract}


\section{Introduction}

Computational models play an essential role across various scientific and engineering disciplines in understanding, simulating, and predicting complex systems. As these models aim for increased fidelity in representing real-world systems, the simulations become more intricate, demanding higher computational resources. 
Surrogate models, also known as metamodels, are widely employed as computationally cheaper approximations of the expensive model. Surrogate models utilize a limited set of evaluations from the original model to construct an efficient, simplified, yet accurate representation of the original model. These models are mainly used in applications that include prediction, uncertainty quantification, sensitivity analysis, and surrogate-assisted optimization \cite{kudela2022recent}. 
However, to accurately represent the complex model, it is necessary to have a sufficient number of model evaluations, which can sometimes be prohibitively expensive. Hence, it is essential to construct efficient experimental designs by developing strategies to minimize the number of deterministic evaluations, which has been an active area of research \cite{liu2018survey,fuhg2021state,dutta2020design}.
Furthermore, ensuring the surrogate model adheres to the constraints of the original model to provide realistic predictions is also crucial in many applications \cite{sharma2024learning,sun2020surrogate}. Addressing these goals holds significant interest within the surrogate modeling community.

Several surrogate modeling methods are available in the literature, with popular ones including polynomial chaos expansions (\PCE{}) \cite{xiu2002wiener, lüthen2021sparse, ghanem2017polynomial}, Gaussian process regression (GPR) \cite{gramacy2020surrogates, rasmussen2006gaussian}, and deep neural networks \cite{lieu2022adaptive, olivier2021bayesian}. Among these methods, \PCE{} is a widely adopted method in the field of uncertainty quantification (\UQ{}). It is employed in the context of stochastic systems where input parameters are subject to variability or randomness and is often preferred for low-to-medium dimensional problems. A
\PCE{} surrogate model approximates the response of a stochastic computational model by a spectral expansion of orthogonal multivariate polynomials with deterministic coefficients. The orthogonal polynomials are selected based on the distributions of the input random variables based on the Weiner-Askey scheme \cite{xiu2002wiener} and can also be constructed based on the arbitrary distributions of the input data \cite{soize2004physical}. Several approaches exist for computing the \PCE{} coefficients, including intrusive approaches such as the stochastic Galerkin (SG) method \cite{ghanem2003stochastic}, which minimizes the error of a finite-order \PCE{} by Galerkin projection onto each polynomial basis to yield a large coupled set of deterministic equations. This approach is called \emph{intrusive} since it requires modification of the original model or simulation code to incorporate the stochastic parameters, making it less flexible in many applications \cite{debusschere2004numerical}. On the other hand, \emph{non-intrusive} approaches use simulations as black boxes, and the calculation of \PCE{} coefficients is based on a set of simulation responses evaluated on the prescribed nodes or collocation points in the stochastic space. This approach to solving \PCE{} coefficients is broadly known as the stochastic collocation method \cite{xiu2005high}, often characterized by three main types: interpolation \cite{narayan2012stochastic}, regression \cite{berveiller2006stochastic}, and pseudo projection method \cite{xiu2007efficient}. A survey of these methods can be found in \cite{xiu2016stochastic}. Of these methods, linear regression is of particular importance for this work, where the \PCE{} coefficients are computed by minimizing the least squares error between the original model response and the \PCE{} approximation at the collocation points. This method is easy to implement and can be coupled with many popular sparse regression algorithms like least angle regression (LAR) \cite{efron2004least, blatman2011adaptive}, Least Absolute Shrinkage and Selection Operator (LASSO) \cite{tibshirani1996regression,zhang2021sparse}, and others. Once the \PCE{} coefficients are computed, the surrogate model can be used to derive estimates of various response statistics, such as its moments or its sensitivity to different input random variables, in a computationally efficient way \cite{sudret2008global, novak2022distribution}. 

\PCE{} has also been recently used as a machine learning (ML) surrogate model for predictions in purely data-driven settings \cite{torre2019data}, demonstrating comparable accuracy to other ML regression models, such as neural networks and support vector machines, without relying on fine-tuning critical hyperparameters and requiring smaller training datasets. In the ML context, the computational model may not be present, and the \PCE{} is utilized to establish a mapping between input-output based on the available training data. The \PCE{} coefficients are computed using linear regression through LAR \cite{blatman2011adaptive}. It has been shown that \PCE{} not only provides accurate pointwise predictions but also output statistics through proper probabilistic characterization of input uncertainties using marginal distributions and copulas. Furthermore, the \PCE{} surrogate model has been demonstrated to be robust to noise in the training dataset \cite{torre2019data}.

In recent years, there has been a notable surge in research interest towards integrating fundamental physical laws and domain knowledge in the training procedure of ML surrogate models to solve problems characterized by a limited dataset and a partial understanding of the underlying physics. This new learning philosophy is referred to as physics-informed machine learning (PIML) or, more generally, as scientific machine learning (\PIML{}) \cite{karniadakis2021physics,cuomo2022scientific}.
\PIML{} offers primarily two advantages over its conventional ML counterparts: better generalization performance with accurate and physically realistic predictions and lower training data requirements. Perhaps the most well-known of this class of methods are the physics-informed neural networks (PINNs), which have had a significant impact across different fields in a relatively short period. Raissi et al. \cite{raissi2019physics} introduced and illustrated the PINNs approach for solving forward and inverse problems involving non-linear partial differential equations (PDEs), framing the problem as an optimization task for minimizing a loss function. PINNs are essentially a mesh-free surrogate model primarily employed for solving governing PDEs. The popularity of PINNs can be attributed 
to their efficiency in solving PDEs in domains with complicated geometries or in very high dimensions that are very difficult to solve numerically \cite{cuomo2022scientific,hu2023tackling}. However, PINNs lack built-in uncertainty quantification capabilities, limiting their applications, especially in parametric uncertainties and noisy data scenarios. Nonetheless, there have been significant research efforts to incorporate \UQ{} capabilities in PINNs. Recently, Zhang et al.~\cite{zhang2019quantifying} combined PINNs with arbitrary polynomial chaos (aPC) to quantify parametric and data uncertainty. 
Yang et al.~\cite{yang2021b} proposed Bayesian PINNs to address aleatoric uncertainty associated with noisy data. Zou et al.~\cite{zou2023correcting} quantify model uncertainty in PINNs. However, there are still significant challenges in incorporating uncertainty in PINNs, and it is an active area of research \cite{psaros2023uncertainty}. 

Another example of \PIML{} is physics-constrained Gaussian process regression (GPR), which incorporates physical constraints or other a priori knowledge into the GPR framework
to supplement limited data and regularize the behavior of the surrogate model \cite{swiler2020survey}. Physics-constrained GPR is a powerful non-parametric Bayesian method that naturally captures the model and data uncertainty while conforming to the underlying physics. This inherent capability leads to improved model accuracy and reliability, particularly in problems with limited or noisy data. Physics-constrained GPR is utilized in many scientific applications to perform \UQ{} of highly complex systems \cite{sharma2024learning}. Similar to PINNs, physics-constrained GPR can be used to solve PDEs; however, its applicability is mainly limited to linear PDEs \cite{raissi2017machine}.  

Building on these methods, we identify \PCE{} as a promising \PIML{} method since it can efficiently handle both ML and \UQ{}-related tasks. However, the idea of incorporating physical constraints in the \PCE{} framework has not been explored in the literature, with our recent work being among the first efforts in this direction \cite{novak2024physics, sharma2023constrained}. We have recently introduced physics-constrained \PCE{} to solve deterministic and stochastic ODEs/PDEs \cite{novak2024physics}. In that work, we extend the linear regression approach to solving \PCE{} coefficients to incorporate known constraints using the method of Lagrange multipliers, which yields a linear system of deterministic equations based on the Karush-Kuhn-Tucker (KKT) stationarity condition. However, this approach is limited to equality-type constraints and is more suited for linear PDEs. 

In the present work, we enhance the capabilities of the novel physics-constrained polynomial chaos expansion (\PC{}) method to handle a broad range of problems in both \PIML{} and \UQ{}. The proposed \PC{} method incorporates various types of known physical constraints, such as governing linear/non-linear PDEs along with associated initial and boundary conditions constraints, inequality-type constraints (e.g., monotonicity, convexity, non-negativity, and others), and other a priori information to perform \PIML{} tasks and leverage the efficient built-in \UQ{} capabilities of the \PCE{} representation. Similarly, the added physics constraints capability improves the uncertainty assessment in \UQ{} related tasks by providing reliable estimates of output uncertainties and reducing the number of expensive computational model evaluations for training. 

While training the \PC{} surrogate model, in addition to evaluating the model at collocation points that constitute the experimental design, we enforce the known constraints at a set of virtual collocation points in both the physical and stochastic domains. This results in solving a constrained least squares optimization problem for the \PC{} coefficients.
The virtual collocation points are different from the collocation points in that they do not require model evaluation. Rather, they employ the \PC{} surrogate model itself to enforce the constraints. 
We further propose a sparse \PC{} by integrating the proposed \PC{} method with Least Angle Regression (LAR)~\cite{efron2004least}, which effectively reduces the number of polynomial basis functions needed to accurately capture the output response. This facilitates the use of the proposed method for high-dimensional problems. 
We demonstrate the effectiveness of the proposed method in handling \PIML{} and \UQ{}-related tasks by applying it to diverse sets of problems, e.g., solving deterministic and stochastic PDEs, performing \UQ{} of a stochastic system with parameters modeled as random fields, and data-driven surrogate modeling of a complex physical system with known physical constraints. 

\section{Methodology}
\label{sec:methodology}
In this section, we present the formulation of the proposed \PC{} by extending the standard \PCE{} framework to incorporate physical constraints. 

\subsection{Polynomial Chaos Expansion}\label{sec:PCE}

\PCE{} is primarily employed for the uncertainty analysis of complex systems represented by expensive computational models, where \UQ{} using Monte Carlo Simulation (MCS) is prohibitively expensive.
Consider a physical system represented by a computational model $\pazocal{M}$ with input random parameter vector 
$\boldsymbol{\xi} = \left(\xi_{1}, \xi_{2}, \ldots, \xi_{M}\right) \in \mathcal{D}_{\boldsymbol{\xi}}\subset \mathbb{R}^M$ having prescribed 
marginal probability density functions (PDFs) $\left\{f_{\xi_i},\ i=1, \ldots, M\right\}$. Due to the randomness of the input, following the Doob-Dynkin Lemma, the scalar output of the model, denoted $Y=\pazocal{M}(\boldsymbol{\xi})$, is also a random variable.
Under the assumption that the output random variable $Y$ has finite variance, it can be represented by a \PCE{} as follows:
\begin{equation}\label{eq:PCE}
Y\equiv\pazocal{M}(\boldsymbol{\xi})=\sum_{\boldsymbol{\alpha} \in \mathbb{N}^M} y_{\boldsymbol{\alpha}} \Psi_{\boldsymbol{\alpha}}(\boldsymbol{\xi}),
\end{equation}
where $y_{\boldsymbol{\alpha}} \in \mathbb{R}$ are the expansion coefficients to be determined, $\Psi_{\boldsymbol{\alpha}}(\boldsymbol{\xi})$ are multivariate polynomials, and $\boldsymbol{\alpha} \in \mathbb{N}^M$ is a multi-index $\boldsymbol{\alpha}=\left(\alpha_1, \ldots, \alpha_M\right)$ that specifies the degree of the multivariate polynomials $\Psi_\alpha$ in each of the input variables $\xi_i$. 
Assuming independent input random variables, the multivariate polynomials $\Psi_{\boldsymbol{\alpha}}$ are constructed as the tensor product of univariate polynomials orthonormal with respect to the marginal PDF of the corresponding variable, i.e.,
\begin{equation}\label{eq:multivariate_basis}
\Psi_{\boldsymbol{\alpha}}(\boldsymbol{\xi})=\prod_{i=1}^M \phi_{\alpha_i}^{(i)}\left(\xi_i\right),
\end{equation}
with
\begin{equation}
\left\langle \phi_{\alpha_{i}}^{(i)}, \phi_{\beta_{i}}^{(i)} \right\rangle = \int_{\mathcal{D}_{\xi_i}} \phi_{\alpha_{i}}^{(i)}(\omega_{i}) \times \phi_{\beta_{i}}^{(i)}(\omega_i) f_{\xi_i}(\omega_{i}) d \omega_{i} = \delta_{\alpha_{i} \beta_{i}},
\end{equation}

where $\phi_{\alpha_i}^{(i)}$ is an orthonormal polynomial in the $i^{th}$ variable of degree $\alpha_i$ and $\delta_{\alpha_{i} \beta_{i}}$ is the Kronecker delta.
Consequently, the multivariate polynomials $\Psi_{\boldsymbol{\alpha}}$ are orthonormal with respect to the input random vector $\boldsymbol{\xi}$, 
and the expansion can utilize basis functions for common distributions per the Weiner-Askey scheme \cite{xiu2002wiener} or the basis can be constructed numerically \cite{soize2004physical}. 

For practical implementation, the \PCE{} in Eq.~\eqref{eq:PCE} is truncated after a finite number of terms $P$ as,
\begin{equation}
Y_{\mathrm{PC}}=\sum_{\boldsymbol{\alpha} \in \pazocal{A}} y_{\boldsymbol{\alpha}} \cdot \Psi_{\boldsymbol{\alpha}}(\boldsymbol{\xi}),
\end{equation}
where $\pazocal{A}$ is a finite set of multi-indices of cardinality $P$. 
The standard truncation scheme selects all polynomials in the $M$ input variables of total degree not exceeding $p$ such that,
\begin{equation}
\pazocal{A}=\left\{\boldsymbol{\alpha} \in \mathbb{N}^M:{\|\boldsymbol{\alpha}\|}_{1} \leq p\right\} .
\end{equation}

The cardinality of the truncated index set $ \pazocal A^{M,p} $ is given by
\begin{equation}\label{eq.:Cardinality PCE}
 \mathrm{card} \: \pazocal A^{M,p}= \frac{\left( M+p \right)!}{M! \: p!}\equiv P . 
\end{equation}
Other truncation schemes can be employed to reduce cardinality by, for example, reducing the number of interaction terms using methods such as hyperbolic truncation \cite{blatman2011adaptive}. 

With the \PCE{} representation established, the next step is to compute the coefficients, $\boldsymbol{y}=\left\{y_{\boldsymbol{\alpha}}, \boldsymbol{\alpha} \in\pazocal{A}\right\}$.
There are several intrusive and non-intrusive approaches in the literature to solve for the \PCE{} coefficients. In this work, we consider a non-intrusive approach based on linear regression since it converges faster in terms of the number of model evaluations,
as shown in \cite{blatman2009adaptive}. 
In this approach, we can express the exact expansion as the sum of a truncated series and a residual:
\begin{equation}
Y=\sum_{\boldsymbol{\alpha} \in \pazocal{A}} y_{\boldsymbol{\alpha}} \Psi_{\boldsymbol{\alpha}}(\boldsymbol{\xi})+\epsilon,
\end{equation}
where $\epsilon$ represents the error induced by truncation.

The term \emph{non-intrusive} implies that the \PCE{} coefficients are computed using a few deterministic evaluations of the expensive original computational model $\pazocal{M}$ for selected samples of the input random variables, referred to as the experimental design (ED), or in the context of supervised machine learning (ML), a labeled training dataset. In the regression setting, we can minimize the least-squares residual of the polynomial approximation over the ED or training dataset to compute the set of coefficients $\boldsymbol{y}=\left\{y_{\boldsymbol{\alpha}}, \boldsymbol{\alpha} \in \pazocal{A}\right\}$ as
\begin{equation}\label{eq:regression}
\hat{\boldsymbol{y}}=\underset{\boldsymbol{y}_{\boldsymbol{\alpha}} \in \mathbb{R}^P}{\LARGE{\arg \min }}  \frac{1}{N} \sum_{i=1}^N\left(\pazocal{M}\left(\boldsymbol{\xi}^{(i)}\right)-\sum_{\boldsymbol{\alpha} \in \pazocal{A}} y_{\boldsymbol{\alpha}} \Psi_{\boldsymbol{\alpha}}\left(\boldsymbol{\xi}^{(i)}\right)\right)^2 ,
\end{equation}
where $N$ is the number of deterministic model evaluations. 
The ordinary least-square (OLS) solution of Eq. \eqref{eq:regression} reads:
\begin{equation}\label{eq:OLS}
\hat{\boldsymbol{y}}=\left(\boldsymbol{A}^{\top} \boldsymbol{A}\right)^{-1} \boldsymbol{A}^{\top} \pazocal{Y},
\end{equation}
where
$$
\boldsymbol{A}=\left\{A_{i j}=\Psi_j\left(\boldsymbol{\xi}^{(i)}\right), i=1, \ldots, N ; j=0, \ldots, P\right\}\ \mathrm{and}\ \pazocal{Y}=\left[\pazocal{M}\left(\boldsymbol{\xi}^{(0)}\right), \ldots, \pazocal{M}\left(\boldsymbol{\xi}^{(N)}\right)\right]^{\top}.
$$
Here, $\boldsymbol{A}$ is called the model design matrix that contains the values of all the polynomial basis functions evaluated at the ED points, and $\pazocal{Y}$ is the model evaluations vector. It was shown that the OLS solution requires at least $\mathcal{O}( P \, \ln (P))$ samples for a stable solution \cite{CohenOptimalWLS}. In practice, the number of model evaluations is typically chosen as $N = kP$, where $k\in[2,3]$.
For $N<P$, the solution of Eq. \eqref{eq:OLS} is no longer unique.
As the polynomial degree ($p$) or input dimensionality ($M$) increases, the number of coefficients increases drastically (see Eq. \eqref{eq.:Cardinality PCE}). Consequently, a large number of model evaluations are necessary to achieve a satisfactory level of accuracy, which becomes prohibitively expensive for costly computational models. This problem is addressed by building an adaptive sparse \PCE{} based on least angle regression (LAR) \cite{blatman2011adaptive}, which effectively reduces the number of polynomial basis functions ($P$) and hence the number of model evaluations.

\subsection{Physically Constrained Polynomial Chaos Expansion}

In this section, we incorporate physical constraints in the \PCE{} regression framework as described in Section \ref{sec:PCE}. We referred to this approach as the physics-constrained polynomial chaos (\PC{}) expansion. Integrating additional knowledge of the computational model through physical constraints enriches the experimental design (ED), thereby considerably reducing the number of expensive computational model evaluations necessary to perform \UQ{}. In ML, this translates into having a relatively smaller training dataset to achieve the desired pointwise accuracy. Further, adding constraints ensures that the \PCE{} surrogate model provides physically realistic predictions across the entire input domain.

The \PC{} method is capable of both \PIML{} and \UQ{}-related tasks without much alteration in its formulation. Without a loss of generality, we define the input vector as
$\boldsymbol{X} = \left[\pmb{\pazocal{X}},\ \boldsymbol{\xi}\right]^T$, where
$\pmb{\pazocal{X}} = \left(x_1, x_2, \ldots, x_n\right) \in \mathcal{D}_{\pmb{\pazocal{X}}}\subset \mathbb{R}^n$ and 
$\boldsymbol{\xi} = \left(\xi_{1}, \xi_{2}, \ldots, \xi_{M}\right) \in \mathcal{D}_{\boldsymbol{\xi}}\subset \mathbb{R}^M$.
The input vector consists of deterministic physical variables $(\pmb{\pazocal{X}})$ and a random vector of parameters $(\boldsymbol{\xi})$. 
In the \PC{} framework, the physical variables $(\pmb{\pazocal{X}})$ play a crucial role in accomplishing the \PIML{} task by enforcing constraints pointwise, thereby regularizing the behavior of the polynomial approximation, which leads to improved prediction accuracy and better generalization error in the physical domain. 
This is particularly necessary because the physical constraints are usually formulated in terms of the physical variables $(\pmb{\pazocal{X}})$ and are not, in general, expressed in terms of the random variables contained in $\boldsymbol{\xi}$. To incorporate $\pmb{\pazocal{X}}$ as input in the \PCE{} representation, we need to assume a distribution (here we assume a uniform distribution) so that we can ensure orthogonality of the polynomial basis either through the Wiener-Askey scheme \cite{xiu2002wiener} or numerically~\cite{soize2004physical}. 
Furthermore, we can easily filter out the influence of the assumed distribution for the physical variables by post-processing the \PC{} coefficients, which will be explained in Section \ref{sec:reduced pce}. 

The random input vector $\boldsymbol{\xi}$ captures the uncertainty in the physical system, considering random parameters, as observed in solving stochastic PDEs. This construction of the input vector for the \PCE{} representation allows it to seamlessly integrate \PIML{} into \UQ{} and vice versa. For example, in the proposed \PC{}, integrating uncertain parameters in \PIML{} problems is straightforward, facilitating \UQ{}. Likewise, incorporating physics information as in \PIML{}, enhances the accuracy of uncertainty assessment in \UQ{} problems.

Next, we describe the training procedure of the \PC{} surrogate model. First, let us consider the deterministic output vector $\boldsymbol{y}=\pazocal{M}(\pmb{\mathcal{X})}$, where $\pmb{\mathcal{X}}= \{\pmb{\pazocal{X}}^{(i)}\}_{i=1}^{n_t}$, $n_t$ represents the number of randomly selected grid points
from the discretized physical domain of $\pazocal{M}$. By incorporating random variables in this model, the output $\boldsymbol{y}$ will differ for each model evaluation based on the input realization of the input random vector $\boldsymbol{\xi}$. Hence, the output is also a random vector given as 
$\boldsymbol{Y}^{(j)} = \pazocal{M}^{(j)}(\pmb{\mathcal{X}},\ \boldsymbol{\xi}^{(j)})$, for $j=1\ \text{to}\ N$, where $N$ is the number of model evaluations corresponding to an experimental design ED. Thus, the training dataset consists of $N \times n_t$ points, where the $n_t$ points in the physical domain are randomly selected for each model evaluation. 
We incorporate the known physical constraints in the \PCE{} framework by reformulating the least squares optimization problem in Eq.~\eqref{eq:regression}.
However, the difficulty in applying constraints is that it typically calls for a condition to hold globally, which is computationally infeasible. Hence, we approach this problem by relaxing the global requirement and enforcing the constraints only at discrete locations in the input domain, referred to as virtual collocation points. It is important to note that these are different from traditional collocation points used in non-intrusive \PCE{} as we are not evaluating the expensive computation model for these points, but rather using the cheap predictions of the \PC{} surrogate model itself to enforce constraints. 

This leads to a constrained optimization problem for solving \PC{} coefficients, given by,
\begin{equation} \label{eq:PC^2optimization}
\hat{\boldsymbol{y}} = \dfrac{1}{N} 
 \underset{\tilde{\boldsymbol{y}}}{\LARGE{\arg \min }} \mathlarger{\mathlarger{\sum}}_{j=1}^N\ \dfrac{1}{n_t}\Big\| \boldsymbol{Y}^{(j)}- \boldsymbol{Y}^{(j)}_{\mathrm{PC}}(\pmb{\mathcal{X}},\ \boldsymbol{\xi}^{(j)})\Big\|^2,
\end{equation}
subject to
\begin{eqnarray*}
\boldsymbol{\pazocal{G}} \left( Y_{\mathrm{PC}}(\boldsymbol{X}_v^{(i)})\right) = 0 \quad i=1,2, \ldots, N_v,\\
\boldsymbol{\pazocal{H}} \left( Y_{\mathrm{PC}}(\boldsymbol{X}_v^{(i)})\right) \geq 0 \quad i=1,2, \ldots, N_v,
\end{eqnarray*}
where $\|.\|$ is the ${l}^2$ norm, $\hat{\boldsymbol{y}}$ is the \PC{} coefficients vector, and $\boldsymbol{X}_v^{(i)} = (\pmb{\pazocal{X}}_v^{(i)},\ \boldsymbol{\xi}_v^{(i)})$ is a virtual collocation point. $N$, $n_t$, and $N_v$ are the numbers of model evaluation samples, physical domain points, and virtual collocation points, respectively. To sample virtual collocation points, we can adopt any space-filling sampling strategy from the literature \cite{liu2018survey}. In this work, we use Latin Hypercube Sampling (LHS)~\cite{mckay2000comparison}. For solving deterministic problems as a \PIML{} algorithm, such as deterministic PDEs, 
we set $N=1$ and drop the \PCE{} dependency on $\boldsymbol{\xi}$. In this context, $n_t$ represents the number of training observations, which could be from experiments or simulations.   

In \PIML{}, physical constraints are generally categorized in two types: (1) equality types, which are often the residual of governing PDEs of the original computational model; and (2) inequality types constraints like non-negativity, monotonicity, and convexity, where we have partial information about the response of the original model. Here, we denoted the equality type and inequality type constraints with ${\pazocal{G}}$ and $\pazocal{H}$, respectively. We formulate them separately in the following subsections.

\subsubsection{Formulation for equality-type constraints}\label{sec:equality constraints}

For equality constraints, the primary focus is on incorporating PDE constraints along with their associated boundary and initial conditions. Here, we first consider the deterministic PDE case, which can be extended to stochastic PDE in a straightforward manner.

Consider the general PDE given by
\begin{equation}
\pmb{\pazocal{L}}[u(\mathbf{x},\ t)]  =\mathcal{L}\left(u, \frac{\partial u}{\partial t}, \frac{\partial u}{\partial \mathbf{x}}, \frac{\partial^2 u}{\partial t^2}, \frac{\partial^2 u}{\partial \mathbf{x}^2}, \ldots;\ \boldsymbol{\gamma}\right) =f( \mathbf{x},\ t), \quad \ \mathbf{x} \in \mathcal{D},\ t \in[0, T],
\end{equation}
where $\pmb{\pazocal{L}}[\cdot]$ is a general differential operator, $u( \mathbf{x},\ t)$ is the true solution to be found, $\boldsymbol{\gamma}$ denotes a parameter vector, $f(\mathbf{x},\ t)$ is a source or sink term, $t$ is time, $\mathbf{x}=$ $\left(x_1, x_2, \ldots, x_n\right)$ is the spatial vector, and $\mathcal{D} \in \mathbb{R}^n$ denotes the spatial domain.
This general PDE is subject to initial conditions,
\begin{equation}
    \boldsymbol{I}[u(\mathbf{x},\ 0)]=g(\mathbf{x}),
\end{equation}
and boundary conditions,
\begin{equation}
   \boldsymbol{B}\left[u\left(\mathbf{x}_b,\ t \right)\right]=h\left( \mathbf{x}_b, t\right), \quad \mathbf{x}_b \in \partial \mathcal{D},\ t \in[0, T] ,
\end{equation}
where  $ \boldsymbol{I}[\cdot]$ and $\boldsymbol{B}[\cdot]$ are initial and boundary differential operators, respectively, and $\partial \mathcal{D}$ is the boundary of the given domain. 

In \PC{}, we approximate the solution of the PDE through a \PCE{} approximation, i.e., $Y_{\mathrm{PC}}(\mathbf{x},\ t)\approx u(\mathbf{x},\ t)$ and enforce the PDE constraints at a set of virtual collocation points.
An essential characteristic of any \PIML{} method is efficiency in evaluating derivatives with respect to the spatial and temporal coordinates. This is achieved efficiently in \PC{} by taking derivatives of the polynomial representation $Y_{\mathrm{PC}}(\mathbf{x},\ t)$, which can be performed through term-wise derivatives of the polynomial basis functions as follows:
\begin{equation}\label{eq:pce_derivatives}
\dfrac{\partial^n Y_{\mathrm{PC}}(\mathbf{x},\ t)}{\partial x_i^n}=\frac{\partial^n\left[\sum_{\boldsymbol{\alpha} \in \pazocal{A}} y_{\boldsymbol{\alpha}} \Psi_{\boldsymbol{\alpha}}(\tilde{\mathbf{x}},\ \tilde{t})\right]}{\partial \tilde{x}_i^{\ n}} \Delta_{\Gamma}^n=\sum_{\boldsymbol{\alpha} \in \pazocal{A}} y_{\boldsymbol{\alpha}}\frac{\partial^n \Psi_{\boldsymbol{\alpha}}(\tilde{\mathbf{x}},\ \tilde{t})}{\partial \tilde{x}_i^{\ n}} \Delta_{\Gamma}^n, \quad i=1\ldots n+1 \ \text{with}\ x_{n+1} = t,
\end{equation}
where $\Delta_\Gamma$ reflects the scaling of the time-space variable $x_i$ and standardized $\tilde{x}_i$, i.e. $\Delta_\Gamma=2/(x_{max}-x_{min})$ for Legendre polynomials defined on $\tilde{x}_i\in\left[-1,1\right]$ orthonormal to $x_i\sim\pazocal{U}\left[x_{min},x_{max}\right]$. 

Substituting the \PCE{} derivatives into the given PDE, yields the \PC{} constraints as,
\begin{align}
    &\boldsymbol{\pazocal{G}}_{\mathrm{PDE}} \left( Y_{\mathrm{PC}}( \mathbf{x},\ t)\right) = \mathcal{L}\left(Y_{\mathrm{PC}}, \frac{\partial Y_{\mathrm{PC}}}{\partial t}, \frac{\partial Y_{\mathrm{PC}}}{\partial \mathbf{x}}, \frac{\partial^2 Y_{\mathrm{PC}}}{\partial t^2}, \frac{\partial^2 Y_{\mathrm{PC}}}{\partial \mathbf{x}^2}, \ldots;\ \boldsymbol{\gamma}\right)  - f( \mathbf{x},\ t) = 0,\\
    &\boldsymbol{\pazocal{G}}_{\mathrm{IC}} \left( Y_{\mathrm{PC}}( \mathbf{x},\ 0)\right) = \boldsymbol{I}[Y_{\mathrm{PC}}(\mathbf{x},\ 0)]-g(\mathbf{x})=0,\\
    &\boldsymbol{\pazocal{G}}_{\mathrm{BC}} \left( Y_{\mathrm{PC}}( \mathbf{x}_b,\ t)\right)=\boldsymbol{B}\left[Y_{\mathrm{PC}}\left(\mathbf{x}_b,\ t \right)\right] - h\left( \mathbf{x}_b,\ t\right) = 0 .
\end{align}
We enforce these constraints at a set of virtual collocation points as described in Eq.~\eqref{eq:PC^2optimization}. 

An effective approach to solve the constrained optimization problem to obtain the \PC{} coefficients is to transform it into an unconstrained optimization problem using an adaptive weighting scheme by treating the constraints as \emph{soft constraints}. This approach offers a straightforward implementation that is easier to solve compared to constrained optimization with \emph{hard constraints} as originally formulated. Moreover, this alternative formulation can leverage a wide range of well-established and efficient algorithms for solving unconstrained optimization problems. 
For the deterministic case where $N=1$ and $\boldsymbol{X}^{(i)} = \pmb{\pazocal{X}}^{(i)} = \left(\mathbf{x}^{(i)},\ t^{(i)} \right)$, $i=1\ \text{to}\ n_t$, the problem is then formulated as,
\begin{equation}\label{eq:unconstrained_optimization_eq}
\hat{\boldsymbol{y}} = \arg \min _{\tilde{\boldsymbol{y}}}\ L_{\PC{}}(\tilde{\boldsymbol{y}})
\end{equation}
where $L_{\PC{}}$ is the total regularized loss of \PC{} approximation given by
\begin{equation}
L_{\PC{}}(\tilde{\boldsymbol{y}})= \lambda_T L_T(\tilde{\boldsymbol{y}})+\lambda_B L_B(\tilde{\boldsymbol{y}})+\lambda_I L_I(\tilde{\boldsymbol{y}})+\lambda_P L_P(\tilde{\boldsymbol{y}})
\end{equation}
where the components are defined as follows:
\begin{align*}
\text{Training Loss}: \quad L_T &= \frac{1}{n_t}\sum_{i=1}^{n_t}\left(u(\pmb{\pazocal{X}}^{(i)}) - Y_{\mathrm{PC}}(\pmb{\pazocal{X}}^{(i)})\right)^2, \\
\text{PDE Loss}: \quad L_{\mathrm{PDE}} &= \frac{1}{n_v}\sum_{i=1}^{n_v} \left(\boldsymbol{\pazocal{G}}_{\mathrm{PDE}}\left(Y_{\mathrm{PC}}(\pmb{\pazocal{X}}^{(i)})\right)\right)^2, \\
\text{IC Loss}: \quad L_{\mathrm{IC}} &= \frac{1}{n_v^{\text{\scriptsize IC}}}\sum_{i=1}^{n_v^{\text{\scriptsize IC}}} \left(\boldsymbol{\pazocal{G}}_{\mathrm{IC}}\left(Y_{\mathrm{PC}}(\pmb{\pazocal{X}}^{(i)})\right)\right)^2, \\
\text{BC Loss}: \quad L_{\mathrm{BC}} &= \frac{1}{n_v^{\text{\scriptsize BC}}}\sum_{i=1}^{n_v^{\text{\scriptsize BC}}} \left(\boldsymbol{\pazocal{G}}_{\mathrm{BC}}\left(Y_{\mathrm{PC}}(\pmb{\pazocal{X}}^{(i)})\right)\right)^2 .
\end{align*}
where $n_{t}$ represents the total number of training data, which could be observation data (e.g., from experiments) or low-fidelity simulation data (e.g., coarse-mesh FEM simulations), and $n_{v},\ n_v^{\text{\scriptsize IC}},\ n_{v}^{\text{\scriptsize BC}}$ are the numbers of virtual collocation points associated with the domain, boundary, and initial conditions, respectively, whose sum represents the total number of virtual collocation points $N_v$ as given in Eq.~\eqref{eq:PC^2optimization}.

An adaptive scheme \cite{liu2021dual} is then adopted to assign the weights of different losses for each iteration of the training process as
\begin{equation}
    \lambda_i=\frac{L_i}{L_T+L_{\mathrm{PDE}}+L_{\mathrm{IC}}+L_{\mathrm{BC}}},\quad i \in\{T, B, I, P\}.
\end{equation}
The weights are proportionate to the individual losses, implying that the optimization algorithm assigns more significance (higher weights) to components of the loss that have a greater impact on the overall loss. This leads to an effective training process and yields an accurate model approximation.   

In comparison to other \PIML{} methods, the distinctive advantage of \PC{} lies in its direct extension from solving deterministic PDEs to solving stochastic PDEs, as in \UQ{}. When solving stochastic PDEs, the unconstrained optimization problem is formulated as in Eq.~\eqref{eq:unconstrained_optimization_eq} by including the stochastic parameter vector $\boldsymbol{\xi}$ into the \PCE{} representation and considering $N$ model evaluations for the training process. With each model evaluation, $n_t$ points in the physical domain are selected to train the surrogate model. This extension is straightforward and, therefore, not shown explicitly here. 

\subsubsection{Formulation for inequality-type constraints}\label{sec:inequality contraints}

Inequality-type constraints generally restrict the solution space based on partial information about the response of the original computational model, leading to more reliable and interpretable predictions for complex physical systems. These constraints capture certain properties or characteristics that the solution must satisfy, e.g., non-negativity, monotonicity, or convexity, but do not usually represent the complete physics of the problem. 

Inequality constraints can be effectively incorporated into the \PC{} framework by introducing penalty factors for constraint violations. We again transform the constrained optimization problem in Eq.~\eqref{eq:PC^2optimization} into an unconstrained optimization problem as,
\begin{equation}\label{eq:unconstrained_optimization_ineq}
\hat{\boldsymbol{y}} = \arg \min _{\tilde{\boldsymbol{y}}}\ L_{\PC{}}(\tilde{\boldsymbol{y}}) = \underset{\tilde{\boldsymbol{y}}}{\LARGE{\arg \min }} \dfrac{1}{N} \mathlarger{\mathlarger{\sum}}_{j=1}^N\ \dfrac{1}{n_t}\Big\| \boldsymbol{Y}^{(j)}- \boldsymbol{Y}^{(j)}_{\mathrm{PC}}(\pmb{\mathcal{X}},\ \boldsymbol{\xi}^{(j)})\Big\|^2 
+\dfrac{1}{N_{v}} \sum_{\mathrm{i}=1}^{N_v} \lambda_i\left\langle \boldsymbol{\pazocal{H}} \left( Y_{\mathrm{PC}}(\boldsymbol{X}_v^{(i)})\right) \right\rangle^2,
\end{equation}
where
$$\left\langle\boldsymbol{\pazocal{H}} \left( Y_{\mathrm{PC}}(\boldsymbol{X}_v^{(i)})\right)\right\rangle = \begin{cases}0 & \text { if constraint is satisfied at $\boldsymbol{X}_v^{(i)}$} \\ \boldsymbol{\pazocal{H}} \left( Y_{\mathrm{PC}}(\boldsymbol{X}_v^{(i)})\right) & \text { if constraint is violated at $\boldsymbol{X}_v^{(i)}$}\end{cases},$$
 and $\lambda_{i}$ is the user-defined penalty factor. Here, for example
 \[
\boldsymbol{\pazocal{H}} \left( Y_{\mathrm{PC}}(\boldsymbol{X}_v^{(i)})\right) = 
\begin{cases}
    Y_{\mathrm{PC}}(\boldsymbol{X}_v^{(i)}) & \text{for non-negativity constraint} \\
    Y'_{\mathrm{PC}}(\boldsymbol{X}_v^{(i)}) & \text{for monotonicity constraint} \\
    Y''_{\mathrm{PC}}(\boldsymbol{X}_v^{(i)}) & \text{for convexity constraint}
\end{cases},
\]
where again derivatives can be obtained from Eq.~\eqref{eq:pce_derivatives}. In a similar manner, we can define any other types of inequality constraints. 

Note that this formulation includes both space-time variables and random variables for the sake of generality. Since the physical variables are included in the \PCE{} representation, it becomes important to condition the \PCE{} response for these variables when estimating statistics at a specific physical location. This can be efficiently achieved by simple post-processing of \PC{} coefficients and is discussed in the next section.

\subsection{Post-processing of \PC{} coefficients}\label{sec:reduced pce}

Due to the orthogonality of the basis functions, the \PCE{} representation allows for powerful and efficient post-processing of the coefficients to estimate various output statistics such as moments, PDF, and Sobol indices \cite{sudret2008global}.
These output statistics are essential in the uncertainty assessment of stochastic systems.
\PC{} inherits these advantages, while the inclusion of deterministic space-time variables in the \PCE{} representation enables  \PIML{}. 
In \PC{}, since a distribution is assumed for the deterministic variables, it's necessary to filter out the influence of these variables when estimating output statistics. To do so, the \PC{} is conditioned on the deterministic space-time variables or other physical variables to obtain local output statistics. This can be done effectively by treating the physical variables as constant in the \PC{} expansion and using the reduced \PCE{}, recently proposed by Novak~\cite{novak2022distribution} and later adapted for KKT-based \PC{} \cite{novak2024physics}. 

Formally stated, the input vector $\boldsymbol{X} = \left[\pmb{\pazocal{X}},\ \boldsymbol{\xi}\right]^{\top}$ contains $n$ deterministic variables and $M$ random variables. Each element $\boldsymbol{\alpha}$ of the set $\pazocal{A}_{\boldsymbol{X}}$ is a $(n+M)$-tuple that specifies the degree of each variable for the corresponding basis functions $\Psi_{\boldsymbol{\alpha}}\left(\pmb{\pazocal{X},\ \boldsymbol{\xi}}\right)$. The elements $\boldsymbol{\alpha}$ can be partitioned according to the variable types they are assigned to as  $\boldsymbol{\alpha}=\left(\boldsymbol{\alpha_{\pmb{\pazocal{X}}}},  \boldsymbol{\alpha}_{\boldsymbol{\xi}}\right)$ where 
$\boldsymbol{\alpha}\in\pazocal{A}_{\boldsymbol{X}}$. 
In this setting, there are elements in $\pazocal{A}_{\boldsymbol{X}}$ that only differ in $\boldsymbol{\alpha}_{\pmb{\pazocal{X}}}$ and have the same $\boldsymbol{\alpha}_{\boldsymbol{\xi}}$, corresponding to the polynomial basis functions having the same degrees in the random variables, but different degrees for the deterministic variables. To condition over the deterministic variables, consider the set $\pazocal{A}_{\boldsymbol{\xi}}$
that contains all the unique $\boldsymbol{\alpha}_{\boldsymbol{\xi}}$. For each $\boldsymbol{\alpha}_{\boldsymbol{\xi}}\in\pazocal{A}_{\boldsymbol{\xi}}$, we define the conditional set
$\pazocal{T}_{\boldsymbol{\alpha}_{\boldsymbol{\xi}}} =\{\boldsymbol{\alpha}_{\pmb{\pazocal{X}}}:(\boldsymbol{\alpha_{\pmb{\pazocal{X}}}},  \boldsymbol{\alpha}_{\boldsymbol{\xi}}) \in \pazocal{A}_{\boldsymbol{X}}, \boldsymbol{\alpha}_{\boldsymbol{\xi}}\in\pazocal{A}_{\boldsymbol{\xi}} \}$, with $\pazocal{T}_{\boldsymbol{\alpha}_{\boldsymbol{\xi}}} \times \pazocal{A}_{\boldsymbol{\xi}} = \pazocal{A}_{\boldsymbol{X}}$.
This is perhaps best illustrated with a simple example shown in Appendix \ref{sec:Reduced_PCE}. 

Using this partitioning, we can then condition the \PC{} response on given values of the physical variables to obtain the corresponding output statistics. After solving for the coefficients $y_{\boldsymbol{\alpha}}$, we have a \PC{} model of the form
\begin{equation}
    Y_{\mathrm{PC}^2}\left(\pmb{\pazocal{X}},\ \boldsymbol{\xi}\right)=\sum_{\boldsymbol{\alpha} \in \pazocal{A}_{\boldsymbol{X}}} y_{\boldsymbol{\alpha}} \cdot \Psi_{\boldsymbol{\alpha}}(\pmb{\pazocal{X}},\ \boldsymbol{\xi}).   \label{eq:PC^2expansion} 
\end{equation}

Using the tensor product construction of the polynomial basis from Eq.~\eqref{eq:multivariate_basis} and partitioned indices $\left(\boldsymbol{\alpha_{\pmb{\pazocal{X}}}}, \boldsymbol{\alpha}_{\boldsymbol{\xi}}\right)$ we can express $\Psi_{\boldsymbol{\alpha}}(\pmb{\pazocal{X}},\ \boldsymbol{\xi})=
\Psi_{\boldsymbol{\alpha_{\pmb{\pazocal{X}}}}}(\pmb{\pazocal{X}}) \Psi_{\boldsymbol{\alpha_{\boldsymbol{\xi}}}}(\boldsymbol{\xi})$ and rewrite Eq. \eqref{eq:PC^2expansion} as
\begin{equation}
    Y_{\mathrm{PC}^2}(\pmb{\pazocal{X}},\ \boldsymbol{\xi})    =\sum_{(\boldsymbol{\alpha_{\pmb{\pazocal{X}}}},  \boldsymbol{\alpha}_{\boldsymbol{\xi}}) \in \pazocal{A}_{\boldsymbol{X}}} y_{({\boldsymbol{\alpha}_{\pmb{\pazocal{X}}}}, \boldsymbol{\alpha}_{\boldsymbol{\xi}})} \cdot {\Psi}_{\boldsymbol{\alpha_{\pmb{\pazocal{X}}}}}(\pmb{\pazocal{X}}) {\Psi}_{\boldsymbol{\alpha}_{\boldsymbol{\xi}}}(\boldsymbol{\xi}), 
\end{equation}
We can then express the \PC{} model conditioned on the deterministic variables as,
\begin{align}
    Y_{\mathrm{PC}^2}(\boldsymbol{\xi} | \pmb{\pazocal{X}})
    &=\sum_{\boldsymbol{\alpha}_{\boldsymbol{\xi}} \in \pazocal{A}_{\boldsymbol{\xi}}} \left(\sum_{\boldsymbol{\alpha}_{\pmb{\pazocal{X}}} \in \pazocal{T}_{\boldsymbol{\alpha}_{\boldsymbol{\xi}}}} y_{({\boldsymbol{\alpha}_{\pmb{\pazocal{X}}}}, \boldsymbol{\alpha}_{\boldsymbol{\xi}})} \cdot {\Psi}_{\boldsymbol{\alpha}_{\pmb{\pazocal{X}}}}(\pmb{\pazocal{X}}) \right) {\Psi}_{\boldsymbol{\alpha}_{\boldsymbol{\xi}}}(\boldsymbol{\xi}).\label{eq:redPC^2-2}
\end{align}
We can further simplify this expression to get the resulting reduced \PC{} model as 
\begin{equation}
    Y_{\mathrm{PC}^2}(\boldsymbol{\xi} | \pmb{\pazocal{X}})
    =\sum_{\boldsymbol{\alpha}_{\boldsymbol{\xi}} \in \pazocal{A}_{\boldsymbol{\xi}}}  {y}_{\boldsymbol{\alpha}_{\boldsymbol{\xi}}} \left( \pmb{\pazocal{X}}\right){\Psi}_{\boldsymbol{\alpha}_{\boldsymbol{\xi}}}(\boldsymbol{\xi}),\label{eq:redPC^2-3}
\end{equation}
where ${y}_{\boldsymbol{\alpha}_{\boldsymbol{\xi}}}$ are deterministic coefficients that depend on the specific values of $\pmb{\pazocal{X}}$. This provides a stochastic \PC{} expansion for each point in space-time.

The orthogonality of the polynomial basis functions facilitates efficient post-processing of the coefficients to estimate 
moments of the output. The first two moments for a given $\pmb{\pazocal{X}}$ are given as

\begin{equation}
\mathbb{E}[Y_{\PC{}}(\pmb{\pazocal{X}})] = {y}_{(\mathbf{0})}(\pmb{\pazocal{X}}), \quad \mathbb{V}[Y_{\mathrm{PC}^2}(\pmb{\pazocal{X}})] = \sum_{\boldsymbol{\alpha_{\boldsymbol{\xi}}} \in \pazocal{A}_{\boldsymbol{\xi}} \backslash\{(\mathbf{0})\}} {y}_{\boldsymbol{\alpha}_{\boldsymbol{\xi}}}^2(\pmb{\pazocal{X}}).
\end{equation}

Additional output statistics, such as Sobol sensitivity indices \cite{sobol1990sensitivity}, can be analytically derived from the expansion coefficients \cite{sudret2008global}.
The full PDF and the higher-order moments of the output response at each point in space-time can be efficiently estimated using MCS by sampling a sufficient number of realizations of \(\boldsymbol{\xi}\) and evaluating the corresponding computationally inexpensive \PC{} response.

\subsection{Sparse \PC{}} \label{sec:sparsePC^2}
This section presents a sparse implementation of the \PC{} based on the popular least-angle regression (LAR) algorithm \cite{efron2004least}. LAR is an efficient algorithm in ML primarily used for feature selection and is particularly well-suited for high-dimensional datasets where the number of features is much larger than the number of observations. In the context of \PC{}, we apply LAR to select those polynomial basis functions (i.e., features) that have the most impact on the model response $Y \equiv \pazocal{M}(\boldsymbol{X})$, among a possibly large set of candidates basis functions ($\pazocal{A}$). 

Before proceeding with the implementation, it is crucial to highlight a few important characteristics of \PC{}:
\begin{enumerate}
    \item \PC{} is quite robust to overfitting, which is attributed to the implicit regularization arising from the physical constraints, similar to other \PIML{} methods such as PINNs. 
    
    \item Incorporating physical constraints in the \PCE{} formulation significantly reduces the number of costly model evaluations required to train the surrogate model. This results in substantial savings in computational resources compared to standard \PCE{}.
\end{enumerate}
Due to point 1, the numerical accuracy of the \PC{} approximation is generally observed to increase with an increase in \PCE{} order $p$ without much issue of overfitting, which is a challenge for standard \PCE{}.
The corresponding increase in $P$ (Eq. \eqref{eq.:Cardinality PCE}) should typically necessitate a larger ED to estimate the coefficients.
However, this is alleviated in \PC{} by incorporating physical constraints (point 2). Nonetheless, increasing $p$ for problems with high input dimensionality leads to a dramatic increase in the number of \PC{} coefficients, which substantially increases the computational cost to solve the optimization problem (Eq. \eqref{eq:PC^2optimization}. Hence, a sparse \PC{} implementation is necessary. 

The LAR procedure applied to the polynomial basis is described below:
\begin{enumerate}
    \item  Input $\boldsymbol{X}$, $\boldsymbol{Y}$, and basis vectors based on the \PCE{} order $p$ as $\left\{\boldsymbol{\Psi}_{\boldsymbol{\alpha}_i}(\boldsymbol{X}), i=0, \ldots, P-1\right\}$.  Initialize the coefficiects as $y_{\boldsymbol{\alpha}_0}, \ldots, y_{\boldsymbol{\alpha}_{P-1}}=0$ and the corresponding predicted response vector $\boldsymbol{Y}_{\mathrm{PC}}=\mathbf{0}$. Define the residual $\boldsymbol{R}=\boldsymbol{Y}-\boldsymbol{Y}_\mathrm{PC}$.

    \item Find the vector $\boldsymbol{\Psi}_{\alpha_j}$ that is most correlated with $\boldsymbol{R}$ and set the truncation set $\pazocal{A}^{(1)} = \left\{\alpha_j\right\}$

    \item Move $y_{\boldsymbol{\alpha}_j}$, from 0 towards the value $\boldsymbol{\Psi}_{\boldsymbol{\alpha}_j}^{\top} \boldsymbol{R}$, until some other basis vector $\boldsymbol{\Psi}_{\boldsymbol{\alpha}_k}$ achieves the same correlation with the current residual as $\boldsymbol{\Psi}_{\boldsymbol{\alpha}_j}$ and set $\pazocal{A}^{(2)} = \left\{\alpha_j,\alpha_k\right\}$

    \item Move jointly $\left\{y_{\alpha_j}, y_{\alpha_k}\right\}^{\top}$ towards their least-squares coefficients of the current residual on $\left\{\boldsymbol{\Psi}_{\alpha_j}, \boldsymbol{\Psi}_{\alpha_k}\right\}$, until some other vector $\boldsymbol{\Psi}_{\alpha_l}$ achieves the same correlation with the current residual and set $\pazocal{A}^{(3)} = \left\{\alpha_j,\alpha_k, \alpha_l\right\}$

    \item Continue this process until all $P$ basis vectors have been included i.e., $\pazocal{A}^{(P)} = \left\{\alpha_j, j=0, \ldots, P-1\right\}$. After $P$ steps, the result will be the complete least-squares solution.

    \item Compute accuracy estimates for each of the $P$ metamodels with truncation sets $\left\{\pazocal{A}^{(1)}, \ldots, \pazocal{A}^{(k)}, \ldots, \pazocal{A}^{(P)}\right\}$ 
    and select the truncation set $\pazocal{A}^{*}$ corresponding to the highest accuracy estimate.
\end{enumerate}

There is an implicit $\pazocal{L}^{1}$ (or Lasso) regularization in LAR that shrinks the coefficients towards zero as more polynomial basis functions are added, eventually yielding a sparse representation containing only a few non-zero coefficients corresponding to the most influential polynomial basis functions.

Consider that after the $k$ steps, the LAR algorithm has included $k$ basis vectors into a truncated index set $\pazocal{A}^{(k)}$. Now, instead of using the associated LAR-based coefficients, one may opt to use the OLS coefficients based on those $k$ predictors. In this setting, LAR is used as a feature selection algorithm and not for estimating the coefficients. This procedure is referred to as LAR–OLS hybrid (or Hybrid LAR), a variant of the original LAR.
As shown in \cite{efron2004least}, the LAR–OLS hybrid will always increase the $R^2$ score (an empirical measure of fit) compared to the original LAR.
Following the same approach, we employ the LAR algorithm to select the most influential $k$ basis functions, which are then utilized to solve for the \PC{} coefficients. 

Blatman et al. \cite{blatman2011adaptive} adapted the LAR procedure to develop an adaptive sparse algorithm for the standard \PCE{}, which we refer to as ``sparse PCE'' in this work. 
In sparse PCE, the critical step in the above LAR procedure is step 6, which requires accuracy estimates for each of the $P$ metamodels to select the ``best'' model. Selection of an appropriate accuracy measure is essential because, for sparse PCE, it must account for the tendency of the model toward overfitting.
This is achieved by employing leave-one-out cross-validation (LOO-CV), where Blatman et al.~\cite{blatman2011adaptive} propose an efficient modified LOO-CV for model selection. 
In contrast, the sparse \PC{} 
implementation is simple and straightforward since it 
is robust to overfitting and provides better generalization error due to the integration of physical constraints.
Therefore, there is typically no need to employ any cross-validation measure to evaluate the performance of the $P$ metamodels obtained from LAR or to implement specific criteria to prevent overfitting.
The general rationale is that the accuracy of \PC{} approximation tends to improve with the addition of more basis functions 
(i.e., increasing $k$ basis functions) to train the \PC{} model. However, the improvement in accuracy is marginal after all the significant polynomial basis functions are included in the truncation set.
Hence, for sparse \PC{}, we aim to select the smallest truncation set that provides the desired level of accuracy.  

\begin{figure}[!ht]
   \centering 
\includegraphics[width=0.6\textwidth]{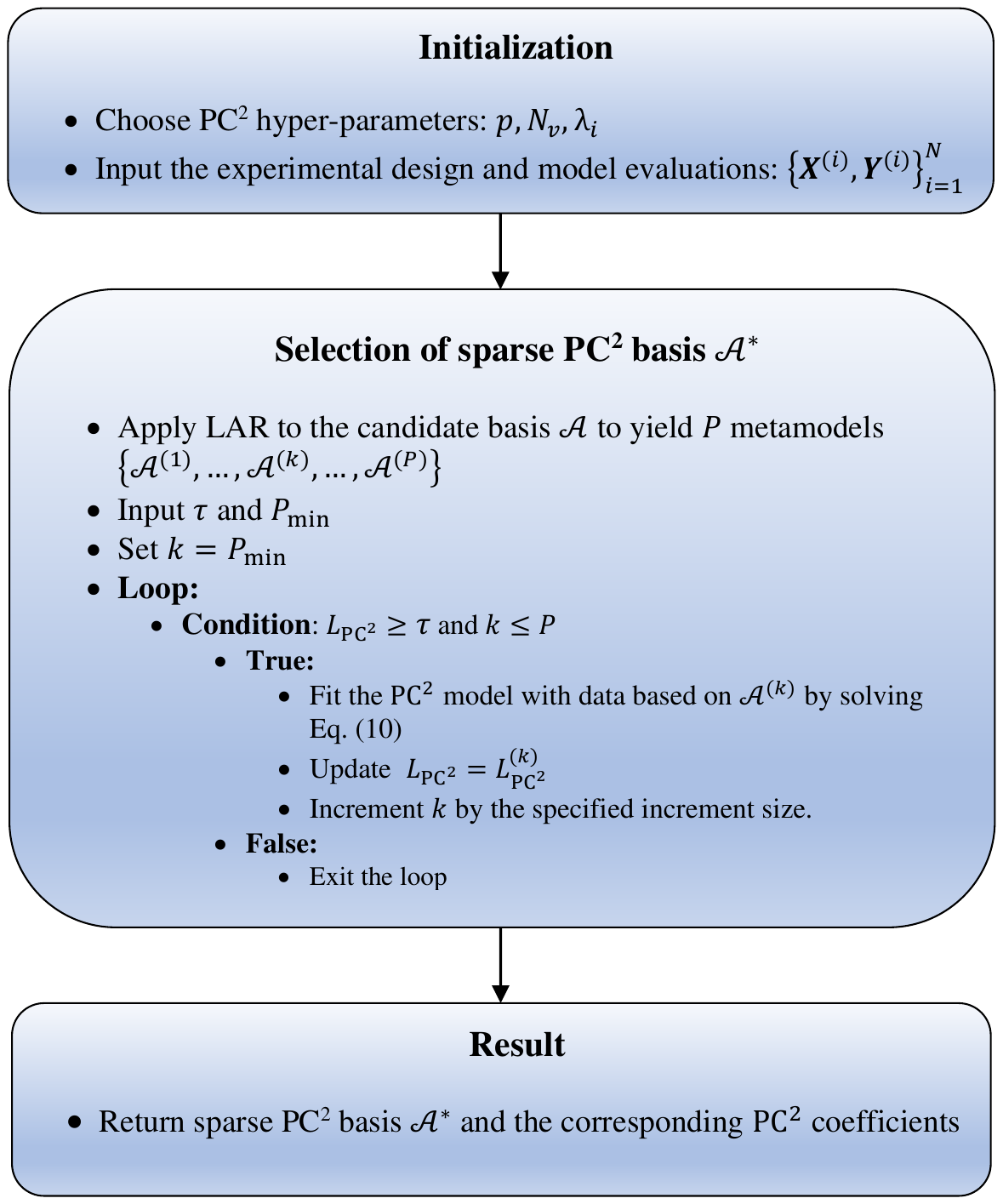}
\caption{Computational flowchart of the sparse \PC{} based on least angle regression (LAR) algorithm.
}
\label{fig:sparse PC^2 flowchart}
\end{figure}

To access the accuracy of the \PC{} approximation, we use the regularized loss $L_{\PC{}}$ (see Eqs.~\eqref{eq:unconstrained_optimization_eq} and \eqref{eq:unconstrained_optimization_ineq}), which contains terms corresponding to training loss and physics-based loss. 
The flowchart of the proposed sparse \PC{} is shown in Figure~\ref{fig:sparse PC^2 flowchart}. The process begins by specifying the tunable \PC{} hyperparameters, specifically the polynomial order $p$, the number of virtual collocation points  $N_{v}$, and the constraint penalty factors $\lambda_i$. 
The value of $p$ can be adjusted based on the user's desired expressivity and maximum dimensionality of the model.
Our empirical findings indicate that selecting a sufficiently large number of virtual collocation points ensures a satisfactory fit. Since the addition of virtual collocation points does not increase computational expense significantly, this number can be selected arbitrarily large, although there is a diminishing return in accuracy as this number grows very large.
We then apply steps 1--5 of the LAR procedure above, which gives $P$ metamodels $\left\{\pazocal{A}^{(1)}, \ldots, \pazocal{A}^{(k)}, \ldots, \pazocal{A}^{(P)}\right\}$ with increasing number of basis functions. Then, rather than applying LOO-CV to assess accuracy, we simply select the smallest basis set that satisfies $L_{\PC{}}<\tau$, where $\tau$ is a user-define error threshold. This is done by iterating over the $P$ metamodels by first initiating $k=\pazocal{A}^{(P_{\text{min}})}$, where $P_{\text{min}}$ denotes the minimum number of basis functions and then incrementing $k$ by a fixed number and solving the constrained optimization problem for the \PC{} coefficients at each iteration until the error threshold is satisfied. 
\section{Numerical results}
In this section, we present numerical examples to demonstrate the capabilities of the proposed physics-constrained polynomial chaos expansion (\PC{}) in both the \PIML{} and \UQ{} settings. 
We apply \PC{} to diverse sets of problems, e.g., solving deterministic and stochastic PDEs,  purely data-driven surrogate modeling of a physical system, and for \UQ{} of a stochastic system with parameters modeled as random fields.
To solve the optimization problem for the \PC{} coefficients (Eqs. \eqref{eq:unconstrained_optimization_eq} and \eqref{eq:unconstrained_optimization_ineq}), we use a quasi-Newton method, BFGS, as implemented in the SciPy library \cite{virtanen2020scipy}. For standard sparse \PCE{}, we use the \UQ{}py software \cite{tsapetis2023uqpy}. For all applications, we use Latin Hypercube Sampling (LHS) \cite{mckay2000comparison} to sample the virtual collocation points.
All the computations are performed on an Apple M1 chip, 8-core CPU, and 8-core GPU with 16 GB RAM.

\subsection{Linear Deterministic and Stochastic PDEs: 2D Heat Equation}

This example aims to highlight the ability of our method to solve linear deterministic and stochastic PDEs. 
Consider the following 2D heat equation with Neumann boundary conditions as
\begin{align}\label{eq:2D_Heat}
\begin{split}
u_t-{\alpha}\left(u_{x x}+u_{y y}\right)&=0, \quad  x, y, t \in[0,1], \\\\
u\left(x, y, 0\right)\ &=0.5\left(\sin (4 \pi x)+\sin (4 \pi y)\right), \\
u_x\left(0, y, t\right)&=0, \\
u_x\left(1, y, t\right)&=0, \\
u_y\left(x, 0, t\right)&=0, \\
u_y\left(x, 1, t\right)&=0,
\end{split}
\end{align}
where $\alpha$ is the thermal diffusivity of the medium and $u$ is the 2D temperature field.

\subsubsection{Deterministic SciML Solution}

We first consider a deterministic case with $\alpha=0.01$ and then extend it to a stochastic case.
To compare the accuracy of the \PC{} solution, we solve this problem numerically using the finite element method (FEM) with the FEniCS library \cite{alnaes2015fenics}.

\begin{figure}[!ht]
   \centering 
    \begin{tabular}{cc}
\includegraphics[width=0.45\textwidth]{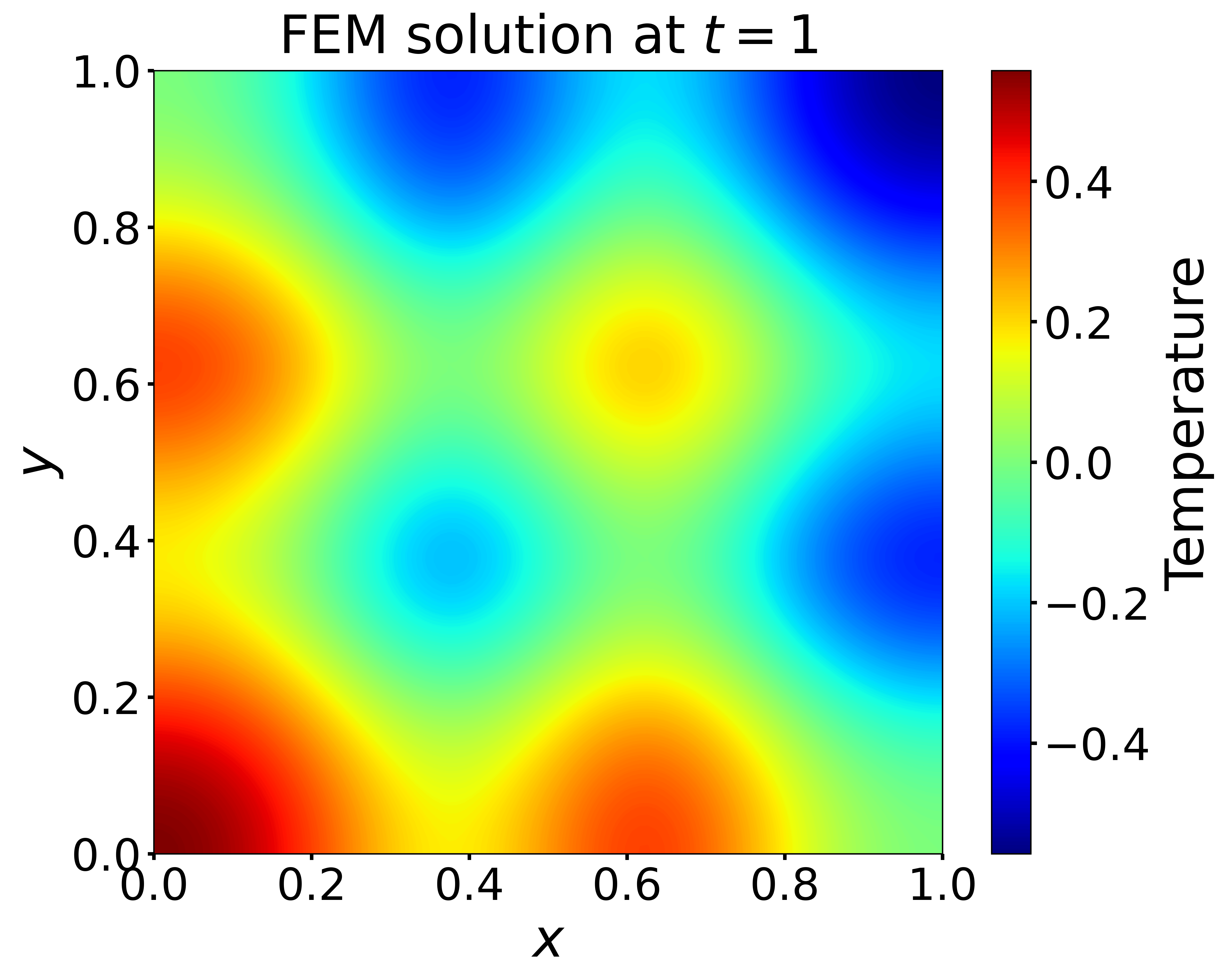} &
\includegraphics[width=0.45\textwidth]{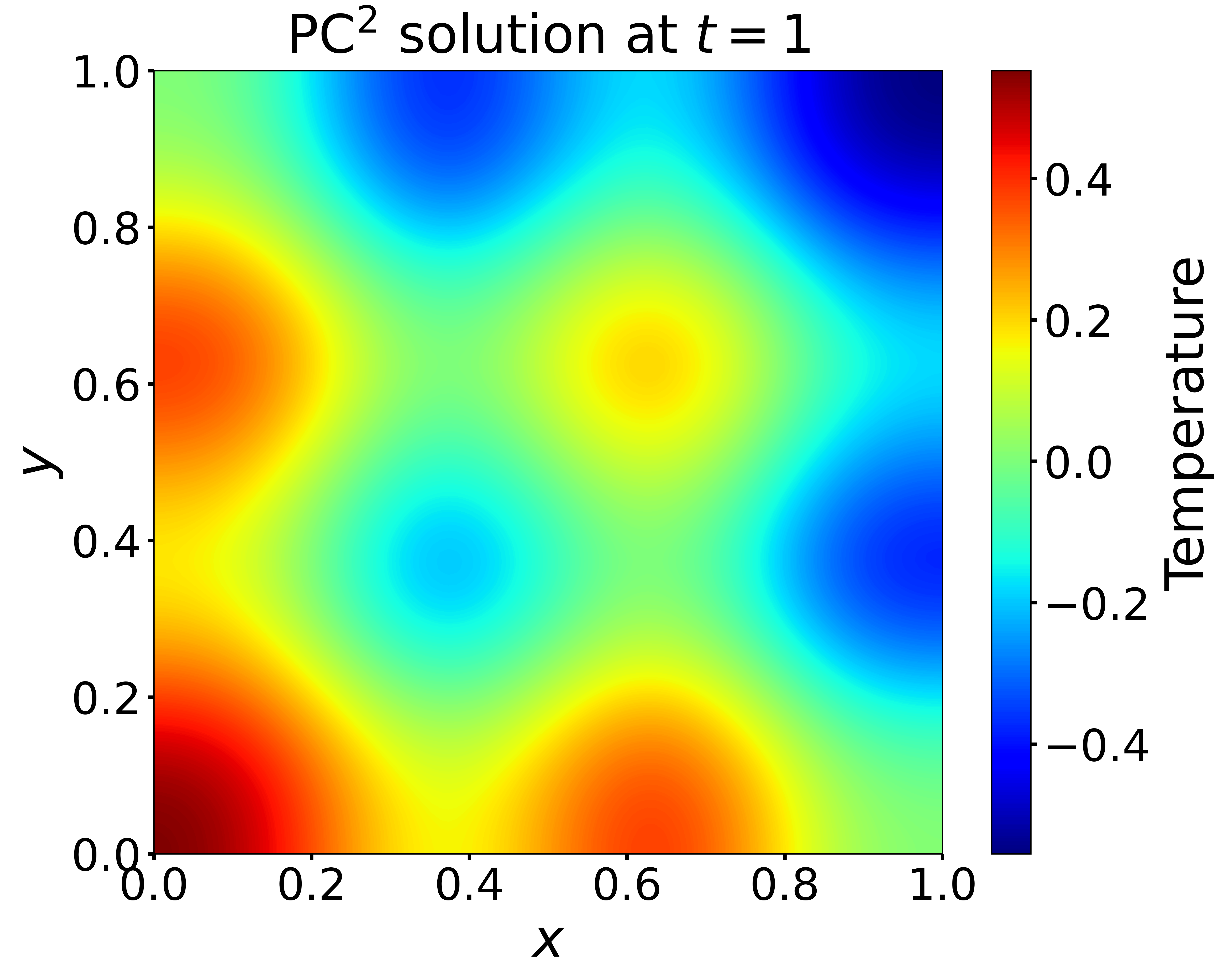} \\
\end{tabular}
\caption{2D Heat Equation: Comparison of the \PC{} solution (right) with the FEM solution (left) at $t=1$, demonstrating very good agreement.}
\label{fig:2D Heat}
\end{figure}

In the context of \PIML{} for solving deterministic PDEs using \PC{}, we can incorporate training data in the form of experimental observations along with the PDE constraints enforced at virtual collocation points. This integration would otherwise be very difficult to achieve in standard PDE solvers, such as FEM. Also, we can incorporate low-fidelity numerical solutions (e.g., coarse-mesh FEM solution) to aid the training process and activate the sparse \PC{} through LAR. 
It is also possible to train \PC{} surrogate model without any labeled training data and enforce the PDE constraints using virtual collocation points, similar to PINNs, where it works directly as a PDE solver.
For this example, we specifically set $n_t=0,\ n_{v}=5000,\ n_v^{\text{\scriptsize IC}}=200,\ n_{v}^{\text{\scriptsize BC}}=200$ with $p=10$ and solve for the full \PC{} model. 
Figure~\ref{fig:2D Heat} shows the \PC{} and the FEM solution at $t=1$, where the \PC{} solution closely matches the FEM solution. 
The MSE evaluated for the entire input domain, discretized as $100^3$ points, is $1.53\times 10^{-04}$, indicating a high level of numerical accuracy in the \PC{} solution.

\begin{figure}[!ht]
{\centerline{\includegraphics[width=0.95\textwidth]{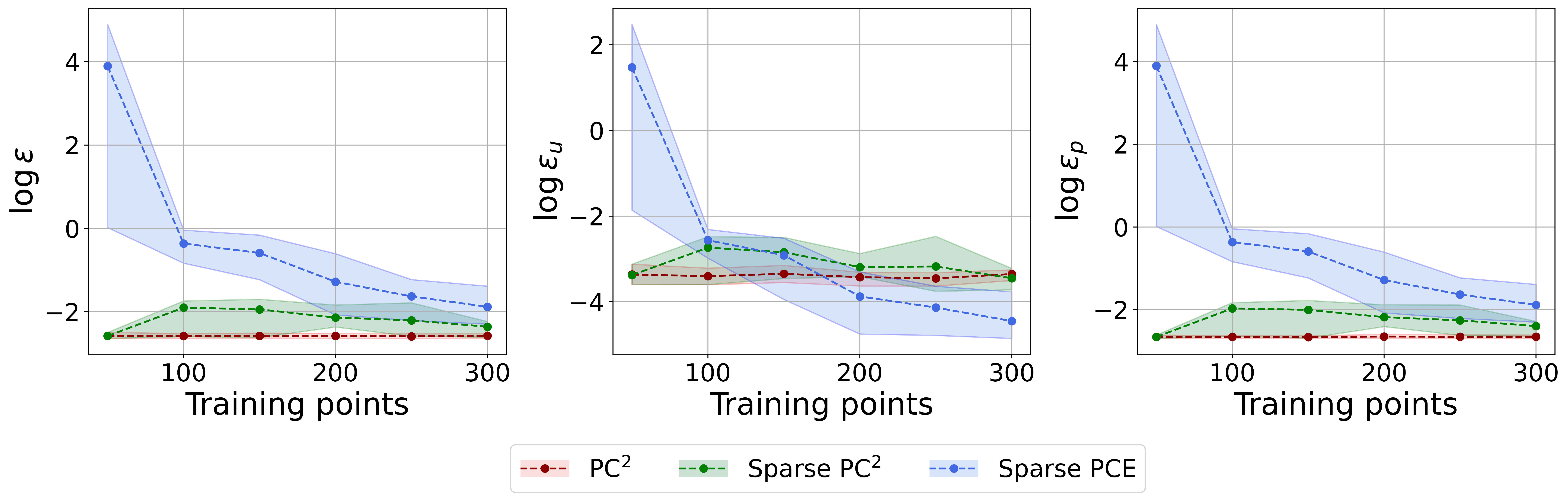}}}
\caption{2D Heat Equation: Convergence plots of the MSEs, $\epsilon=\epsilon_u+\epsilon_p$, for \PC{}, sparse \PC{} and sparse \PCE{} for an increasing number of training points, where
$\epsilon_u$ and $\epsilon_p$ are computed with respect to the FEM solution $u(x,\ t)$ and physics-based residuals associated with the PDE, BCs, and ICs, respectively. 
The dotted line indicates the average relative error, and the shaded area represents the minimum and maximum error across 10 repetitions.}
\label{fig:2D_Heat_convergence}
\end{figure}

Next, we compare the performance of \PC{}, sparse \PC{} 
and standard sparse \PCE{} for increasing training set size
in Figure~\ref{fig:2D_Heat_convergence}. 
Since sparse PCE and sparse \PC{} are based on LAR, we need an experimental design or training data to perform feature selection using the LAR algorithm. For comparison, we consider a modified MSE given by $\epsilon = \epsilon_u + \epsilon_p$, where the total MSE $\epsilon$ is decomposed into the solution error $\epsilon_u$ and the physical constraint error $\epsilon_p$ accounting for the PDE, BC, and IC losses. Errors are computed from a validation set containing $100^3$ grid points across 
the entire domain.
In Figure \ref{fig:2D_Heat_convergence}, the dotted line represents the average error from 10 repetitions, while the shaded area encompasses the range from the minimum to the maximum error observed in the 10 repetitions to represent each method's best and worst performance, respectively. 

\begin{figure}[!ht]
  \centering
  \includegraphics[width=0.5\textwidth]{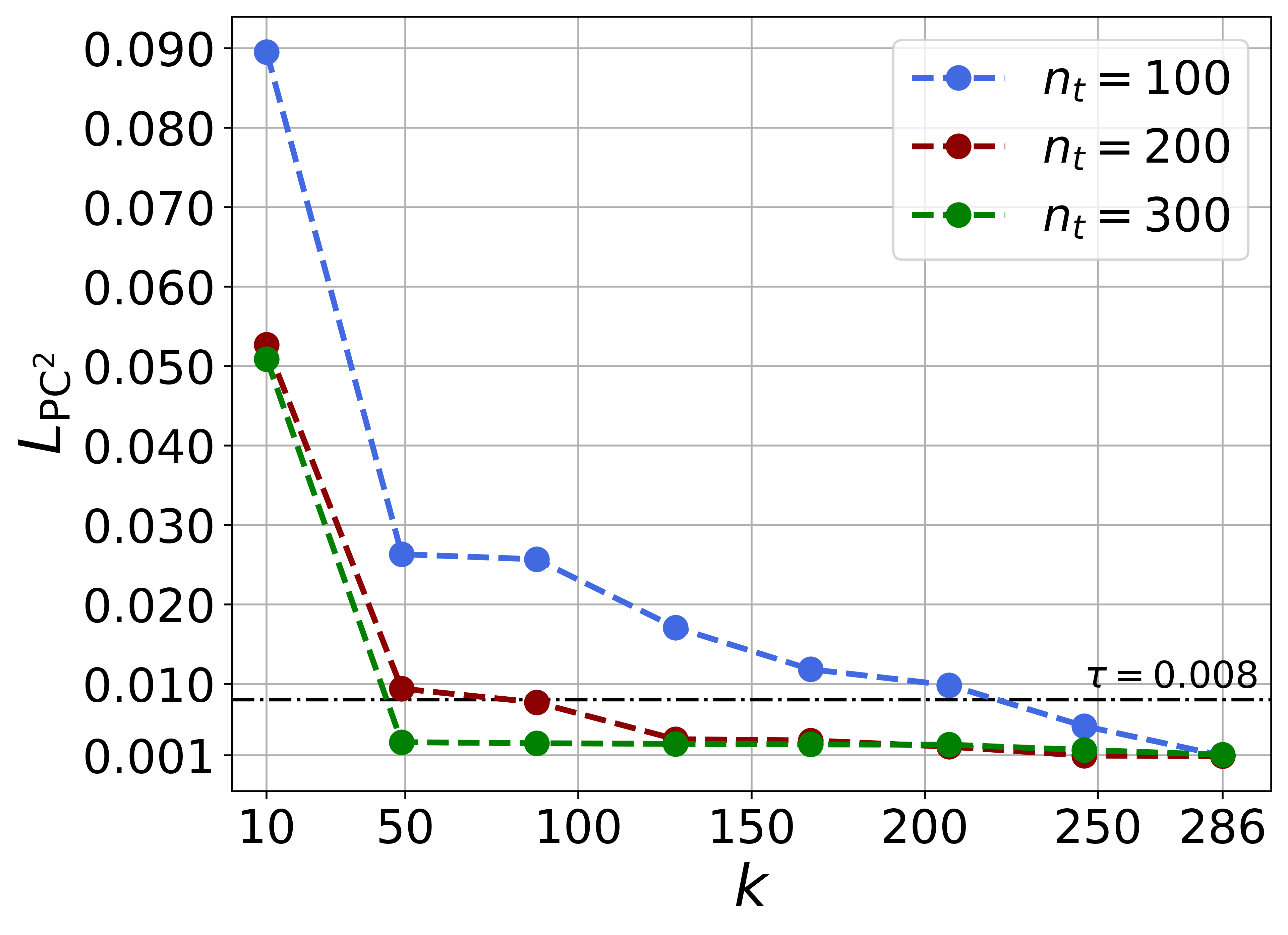}
  \caption{2D Heat Equation: Decay of the regularized loss, $L_\PC{}$, with respect to the number of polynomial basis functions $k$ for different training sample sizes ($n_t$), indicating an improvement in \PC{} sparsity as $n_t$ increases for the user-defined target error, $\tau=0.008$.}
  \label{fig:sparsity}
\end{figure}

As expected, the performance of the full-order \PC{} is not influenced by the number of training points since it already ensures that the given PDE is satisfied at all virtual collocation points in the input domain. The sparse \PC{}, on the other hand, depends on the user-defined target error, $\tau$ (as described in Section \ref{sec:sparsePC^2}), which we set as $\tau=0.008$. As the number of training points increases, the prediction accuracy of the LAR model generally improves, which results in enhanced performance in terms of identifying the most significant polynomial basis functions. This, in turn, improves the sparsity of sparse \PC{}, which is shown in Figure~\ref{fig:sparsity}, where it can be observed that sparse \PC{} requires fewer basis functions as the number of training points increases to achieve the desired error threshold. 
Consequently, it can be observed from Figure \ref{fig:2D_Heat_convergence} that the MSE for both response and physics for sparse \PC{}  is close to the \PC{} for the smaller number of training points, which means it needs a higher number of polynomial basis functions to achieve a given target error. However, as the number of training points increases, the sparsity improves, and sparse \PC{} hits the target error with relatively fewer polynomial basis functions, ultimately converging to the full \PC{} solution.

Finally, both $\epsilon_u$ and $\epsilon_p$ for the sparse PCE decrease as the number of training points increases (Figure~\ref{fig:2D_Heat_convergence}), and it takes around 200 points to achieve comparable $\epsilon_u$ as \PC{} and sparse \PC{}. However, it never achieves the same level of accuracy in the corresponding physics-based MSE, $\epsilon_p$, which suggests that 
for regions in input domains with fewer training data or complicated response surface,
the sparse PCE performs poorly in respecting the physics, leading to an overall increase in the total error $\epsilon$.

Liu and Wang~\cite{liu2021dual} solve this same 2D heat equation with identical initial and boundary conditions using PINNs. Table \ref{tab:2D heat} compares the numerical accuracy and computational efficiency of the proposed full and sparse \PC{} with that of PINNs, as reported in \cite{liu2021dual}. \PC{} provides an order of magnitude better accuracy in a fraction of the computational time without requiring any training samples, in contrast to the 746 samples used for PINNs. With the same training samples as PINNs, the sparse \PC{} just takes 21 seconds to achieve the same level of accuracy as the full \PC{}.  
\begin{table}[h!]
\caption{Comparison of different models in solving 2D Heat equation}\label{3}
\vspace{0.2cm}
\centering
\begin{tabular}{cccc}  \hline
Models     & MSE at $t=1$ & Training time (s) & Training data	\\ 
\hline
PINNs \cite{liu2021dual} & $3.24 \times 10^{-4}$ & 2259 & 746 \\
\PC{} & $6.36 \times 10^{-5}$ & 232 & 0 \\
Sparse \PC{} & $6.42 \times 10^{-5}$ &21 & 746 \\
 \hline
\end{tabular}\label{tab:2D heat}
\end{table}

\subsubsection{Stochastic Solution and Uncertainty Quantification}

Having demonstrated the superior performance of the proposed method for \PIML{} tasks, we now use the proposed method to solve the stochastic PDEs to integrate \UQ{}, which is straightforward with \PC{} in contrast to other \PIML{} methods such as PINNs. 
Here, we consider the stochastic case with parameter $\alpha$ modeled as a uniformly distributed random variable, $\alpha\sim\pazocal{U}[0.001,0.01]$. Again, in contrast to standard \PCE{}, it is possible to use \PC{} to solve this stochastic PDE without any model evaluations. Moreover, we can obtain output statistics at each point in the spatial and temporal domain from a single training, and the trained surrogate model conforms to the underlying PDE constraints over the entire physical and stochastic domains.

\begin{figure}[!ht]
   \centering 
    \begin{tabular}{c}
\includegraphics[width=0.95\textwidth]{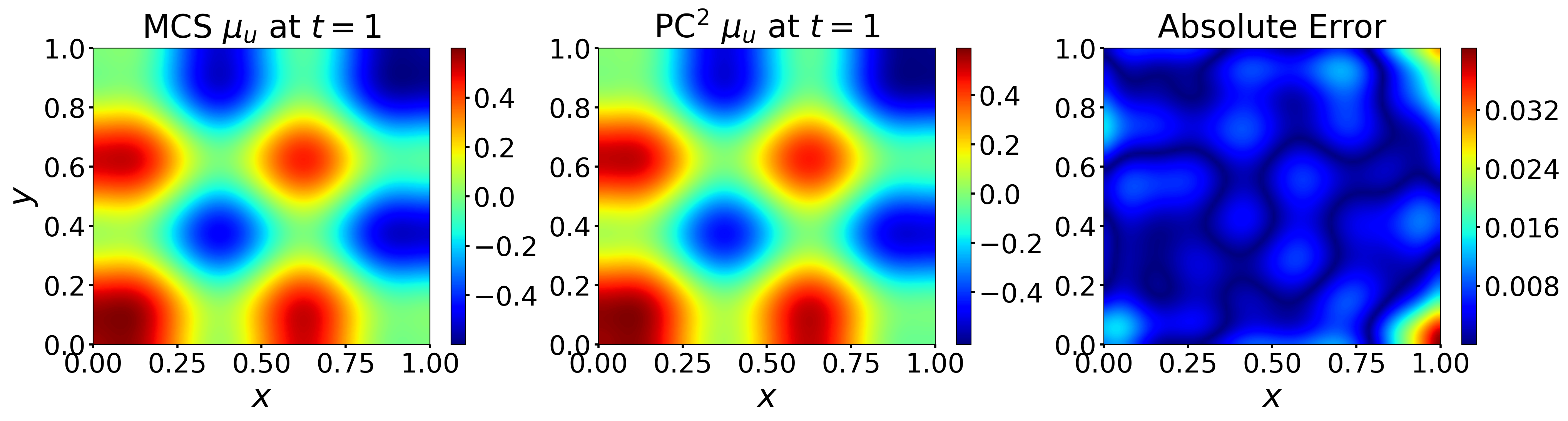}\\
\includegraphics[width=0.95\textwidth]{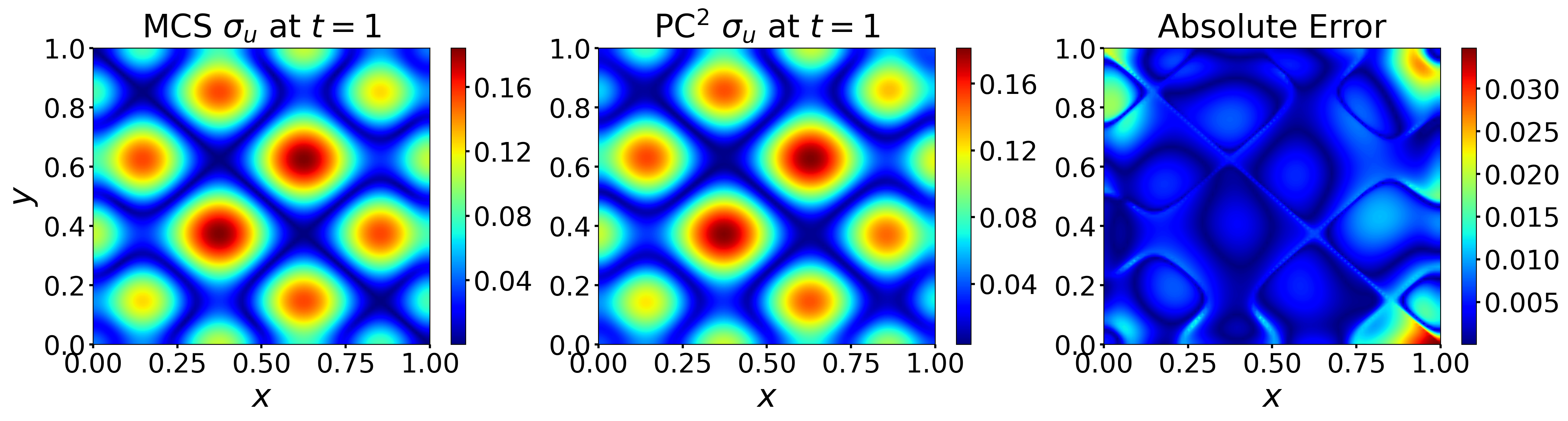}
\end{tabular}
\caption{UQ plots for the stochastic 2D heat equation: \textit{Top panel}: Plots of the mean $\mu_u(x,\ y)$ at $t=1$ obtained by MCS and \PC{}, along with their absolute error. \textit{Bottom panel}: Plots of the standard deviation of $\sigma_u(x,\ y)$ at $t=1$ obtained by MCS and \PC{}, along with their absolute error. 
The plots demonstrate the excellent performance of \PC{} in solving the stochastic 2D heat equation equation without requiring model evaluations.}
\label{fig:2D heat UQ}
\end{figure}

Figure \ref{fig:2D heat UQ} shows the mean, $\mu_u$, and standard deviation, $\sigma_u$, of the solution at $t=1$ using \PC{} ($n_{v}=15000,\ n_v^{\text{\scriptsize IC}}=1000,\ n_{v}^{\text{\scriptsize BC}}=4000$, and $p=10$) and Monte Carlo simulation (MCS), along with the respective absolute error. We use 10,000 model evaluations for MCS to estimate the output statistics. \PC{}, on the other hand, does not require any model evaluations, and the statistics are
efficiently computed by post-processing the obtained \PC{} coefficients (see Section \ref{sec:reduced pce}). From Figure \ref{fig:2D heat UQ}, it is evident that \PC{} provides an accurate approximation of the moments for all the spatial points without needing any model evaluations. 
\begin{table}[!b]
\caption{Comparison of numerical accuracy and computational efficiency of \PC{} and sparse \PC{} in solving stochastic 2D Heat equation with respect to MCS.}
\vspace{0.2cm}
\centering
\begin{tabular}{ccccc}  \hline
Models     & MAE of $\mu_{u}$  & MAE of $\sigma_{u}$ & Model  & Training  	\\ 
& at $t=1$&  at $t=1$&evaluations&time (s)\\
\hline
MCS & - & - & 10000& 56800 \\
\PC{} & 0.004367 & 0.0047222 & 0& 7800 \\
Sparse \PC{} & 0.002308 &0.003726 & 10& 326 \\
 \hline
\end{tabular}\label{tab:2D heat UQ}
\end{table}

Table \ref{tab:2D heat UQ} compares the performance of the proposed full and sparse $\PC{}$ methods with MCS in terms of numerical accuracy and computational time. 
To assess the numerical accuracy, we compute the mean absolute error (MAE) of $\mu_u(x,\ y,\ t=1)$ and $\sigma_u(x,\ y,\ t=1)$ evaluated by averaging the absolute errors at each spatial point. 
\PC{} achieves a high level of accuracy at a fraction of computational time. 
Even though a desirable accuracy can be reached with a lower \PCE{} order ($p$) with significantly less computational time, we deliberately consider a higher order $p=10$ having 1001 polynomial basis functions to compare with sparse \PC{}. This full \PC{} is highly over-parameterized. 
From Table \ref{tab:2D heat UQ}, we can see that considering only a small number of model evaluations activates the sparse \PC{}, which significantly reduces the total polynomial basis. 
Consequently, the computation time is drastically reduced compared to \PC{}. The sparse \PC{} also yields a significant improvement in the numerical accuracy, which results from a much simpler optimization process with fewer coefficients to compute. 

\subsection{Non-linear Deterministic and Stochastic PDEs: Burgers' Equation}
This example highlights the ability of the proposed \PC{} to solve non-linear deterministic and stochastic PDEs. 
Consider the following 1D viscous Burgers' equation, a non-linear PDE describing viscous fluid flow, with Dirichlet boundary conditions as
\begin{align}
\begin{split}
u_t+uu_{x}&=\nu u_{x x}, \quad  x \in[0,1],\  t \in [0,0.3], \\\\
u\left(x,0\right)&=\sin(\pi x), \\
u\left(0, t\right)&=0, \\
u\left(1, t\right)&=0, \\
\end{split}
\end{align}
where $\nu$ is the fluid viscosity and $u$ represents the fluid velocity.

\subsubsection{Deterministic SciML Solution}

We first consider a deterministic case with $\nu=0.001$ where
again, we first solve the equation numerically using FEM with the FEniCS library \cite{alnaes2015fenics} to compare the accuracy of the \PC{} solution. 
For \PC{}, we set $p=20$ and the number of virtual collocation points as  $n_{v}=2000,\ n_v^{\text{\scriptsize IC}}=100,\ n_{v}^{\text{\scriptsize BC}}=100$.  
As in the previous example, no labeled training data is used to train the \PC{} surrogate model.

\begin{figure}[!ht]%
 \centering
 \captionsetup[subfloat]{labelformat=empty} 
 \subfloat[]{\includegraphics[width=0.9\textwidth]{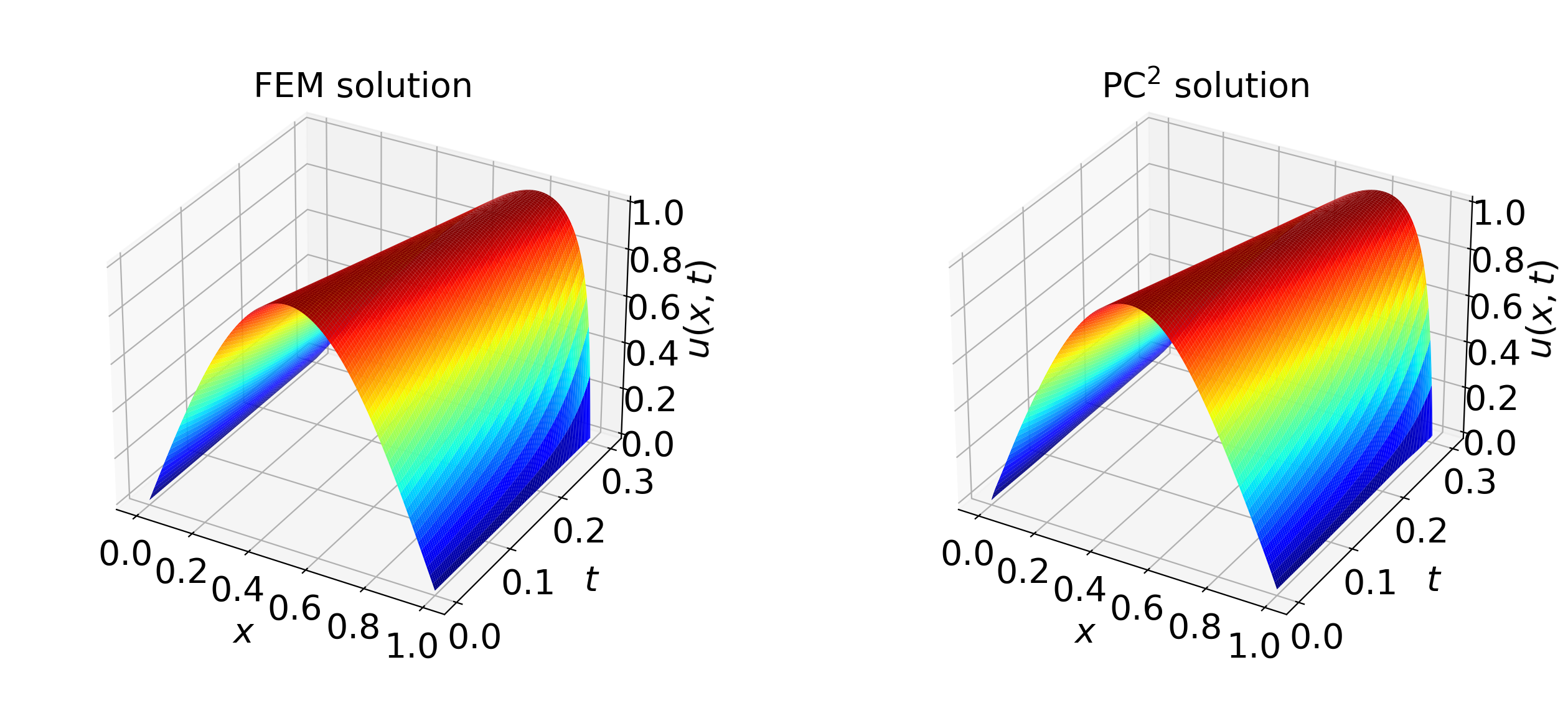}}
 \\
 \subfloat[]{\includegraphics[width=0.9\textwidth]{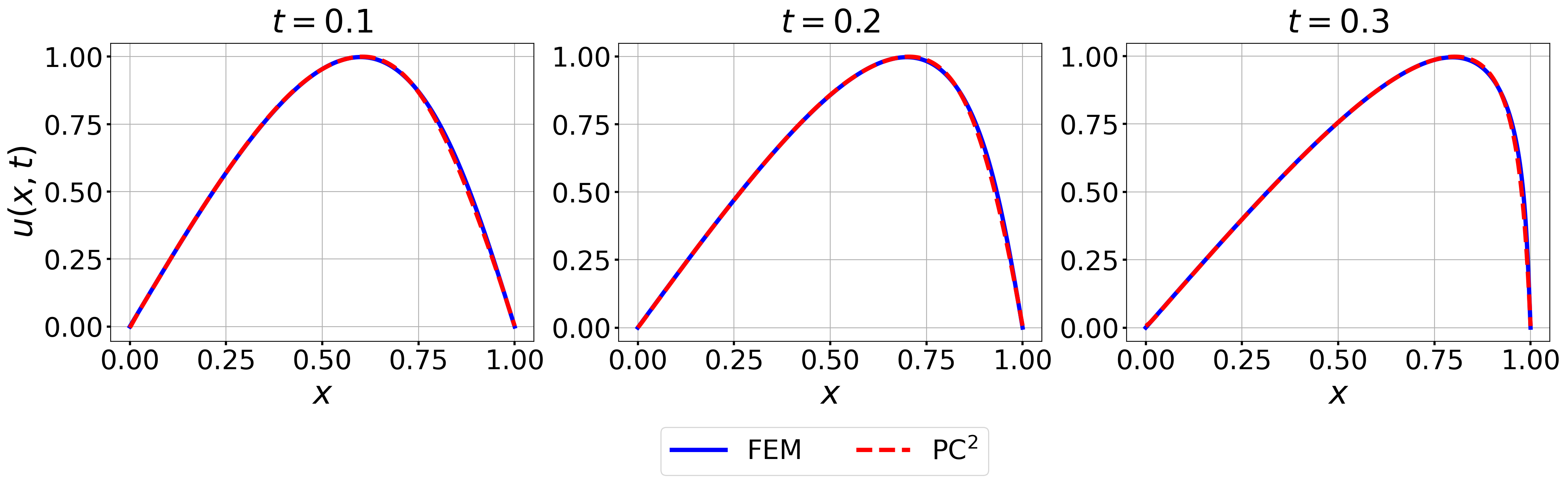}}
 \caption{1D Burgers' Equation: Comparison of the \PC{} solution with the FEM solution demonstrating a very good agreement between the two solutions. \textit{Top panel}: FEM solution and \PC{} over the full space-time domain. \textit{Bottom panel}: FEM and \PC{} solution at specific time instants.} 
 \label{fig:Burgers_Eq_solution}%
\end{figure}

Figure \ref{fig:Burgers_Eq_solution} shows the \PC{} and the FEM solution over the given input domain. It is evident that the \PC{} provides an excellent approximation compared to the FEM solution, which demonstrates the effectiveness of the proposed method in solving non-linear PDEs. Table \ref{tab:Burgers eq} shows the relative numerical error of the full and sparse \PC{} compared to the FEM solution and compares their computational efficiency. The results show excellent numerical accuracy with high computational efficiency for both \PC{} approaches. 

\begin{figure}[!ht]
{\centerline{\includegraphics[width=0.95\textwidth]{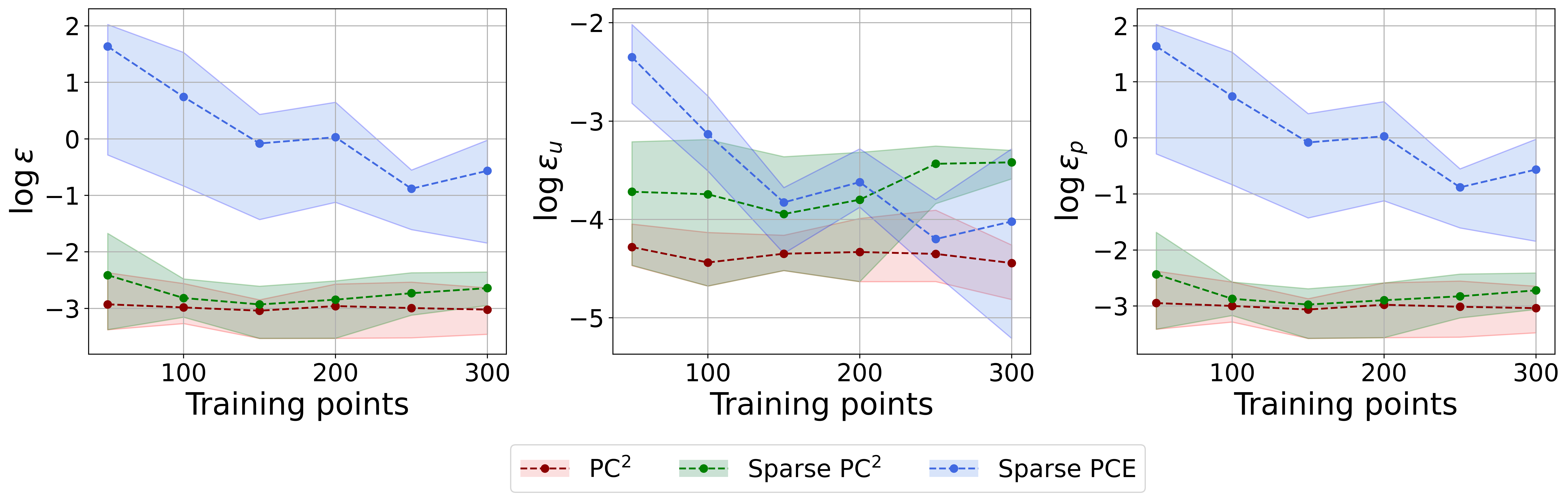}}}
\caption{Burgers' Equation: Convergence plots of the MSEs, $\epsilon=\epsilon_u+\epsilon_p$, for \PC{}, sparse \PC{} and sparse \PCE{} for an increasing number of training points, where
$\epsilon_u$ and $\epsilon_p$ are computed with respect to the FEM solution $u(x,\ t)$ and physics-based residuals associated with the PDE, BCs, and ICs, respectively. 
The dotted line indicates the average relative error and the shaded area represents the minimum and maximum error across 10 repetitions.}

\label{fig:Burgers_convergence}
\end{figure}

Again, to compare the \PC{} with standard sparse PCE, Figure \ref{fig:Burgers_convergence} plots the total MSE, $\epsilon$, and its response and physics-based contributions, $\epsilon_u$ and $\epsilon_p$, for \PC{}, sparse  \PC{}, and sparse PCE for increasing training samples. The errors are computed from a validation set containing $100^2$ grid points across the entire domain.
The convergence trends of the compared methods are similar to the previous example with a notably poor performance by sparse PCE in $\epsilon_p$, 
which is expected since the problem is non-linear and the physics are difficult to adhere to without explicit constraints.
In particular, the sparse PCE is unable to capture the non-linearity as $t$ increases, which results in an overall increase in $\epsilon_p$.
On the other hand, the performance of \PC{} and sparse \PC{} are significantly better, as suggested by the plots of the total MSE, $\epsilon$.
\begin{table}[!ht]
\caption{Comparison of \PC{} and sparse \PC{} models in solving Burgers' equation. MSE computed with respect to the FEM solution.}
\vspace{0.2cm}
\centering
\begin{tabular}{cccc}  \hline
Models     & MSE & Training time (s) & Training data	\\ 
\hline
\PC{} & $4.54 \times 10^{-5}$ &68 & 0 \\
Sparse \PC{} & $3.33 \times 10^{-4}$ &15 & 500 \\
 \hline
\end{tabular}\label{tab:Burgers eq}
\end{table}
This indicates that the proposed method is capable of accurately capturing the complexity of the response $u$ with time. However, with the evolution of the response with time in this problem, the solution becomes non-differentiable. This is difficult to approximate using polynomials. Hence, there is a limit in time $t$ for which \PC{} provides an accurate approximation. There are potential ways to mitigate this issue by adopting, for example, domain adaptive local \PCE{} \cite{novak2023active} for \PC{}, which is a subject of future work. Nevertheless, the performance is significantly superior to the standard \PCE{}.  

\subsubsection{Stochastic Solution and Uncertainty Quantification}

\begin{figure}[!ht]
   \centering 
    \begin{tabular}{c}
\includegraphics[width=0.95\textwidth]{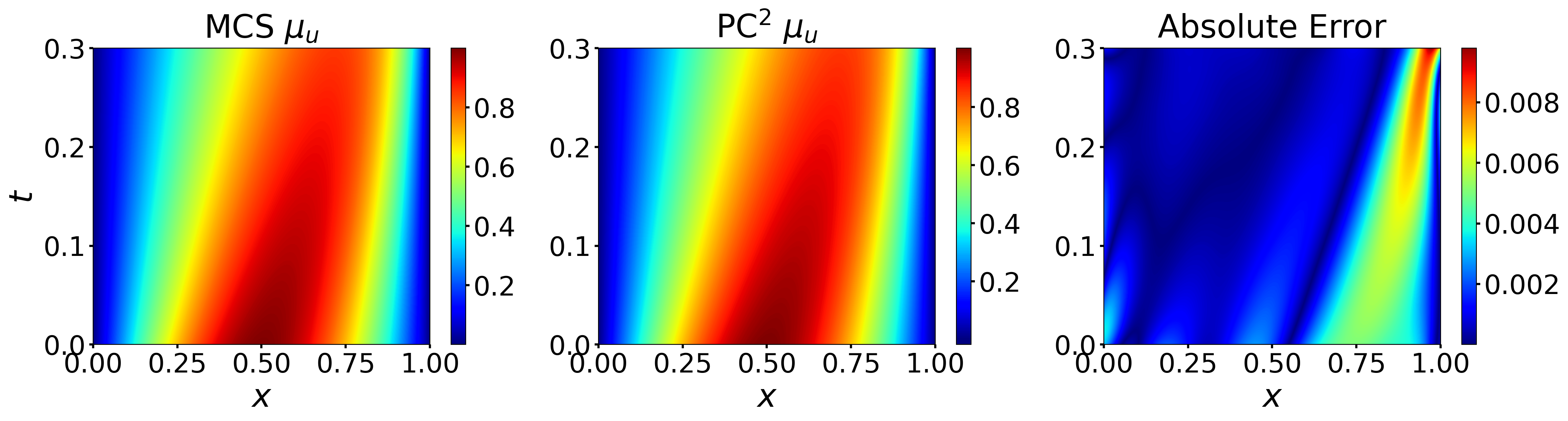}\\
\includegraphics[width=0.95\textwidth]{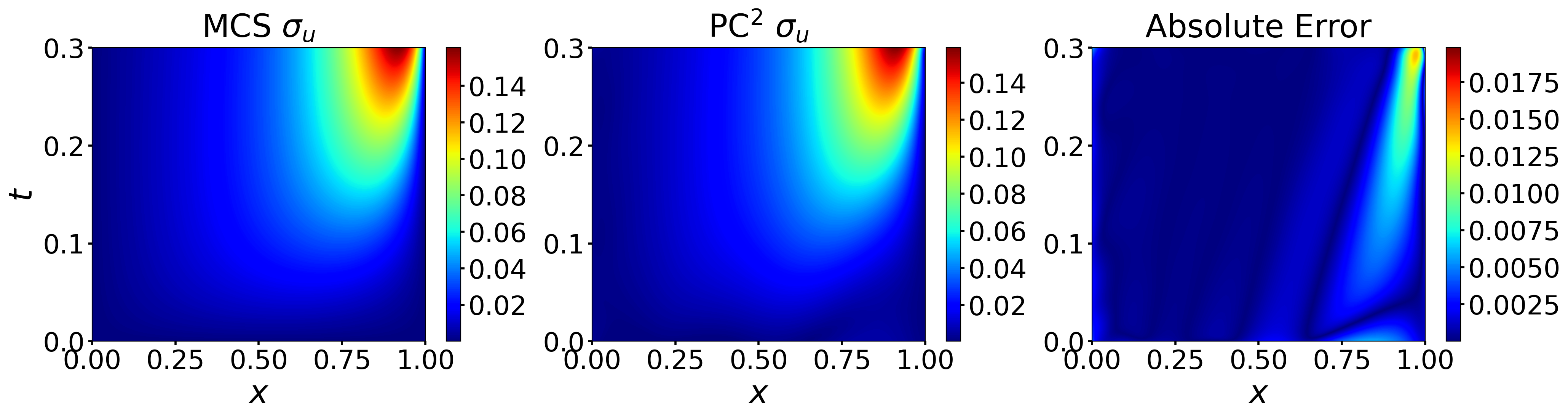}
\end{tabular}
\caption{UQ plots for the 1D stochastic Burgers' equation: \textit{Top panel}: Plots of the mean $\mu_u(x,\ t)$ obtained by MCS and \PC{}, along with their absolute error. \textit{Bottom panel}: Plots of the standard deviation of $\sigma_u(x,\ t)$ obtained by MCS and \PC{}, along with their absolute error. 
The plots demonstrate the excellent performance of \PC{} in solving the 1D stochastic Burgers' equation without requiring model evaluations.}

\label{fig:Burgers UQ}
\end{figure}

Next, we consider the stochastic case with $\nu\sim\pazocal{U}[0.001,0.1]$. Figure \ref{fig:Burgers UQ} shows the mean, $\mu_u(x,t)$, and standard deviation, $\sigma_u(x,t)$ of the solution of the 1D stochastic Burgers' equation using \PC{} and MCS, along with the respective absolute error. Again, we have not used any model evaluations to train the \PC{} surrogate model, in contrast to 10,000 model evaluations for MCS. From Figure \ref{fig:Burgers UQ}, it is evident that \PC{} provides an accurate approximation of the moments.

Table \ref{tab:Burgers UQ} compares the numerical accuracy and computational efficiency of the full-order and sparse \PC{} with MCS. Again, for this example, we considered a high \PCE{} order with $p=18$ (corresponding to 1330 polynomial basis functions) to accurately estimate the output statistics and to compare the \PC{} with the sparse \PC{}. As can be observed from Table \ref{tab:Burgers UQ}, the sparse \PC{} achieves excellent accuracy in estimating the mean and standard deviation compared to MCS with substantial computational time saving by utilizing just a few model evaluations. The full \PC{} is over-parameterized again, and hence it takes considerable training time compared to the sparse \PC{}, but is still significantly less expensive than the MCS.

\begin{table}[!ht]
\caption{Comparison of numerical accuracy and computational efficiency of \PC{} and sparse \PC{} in solving stochastic Burgers' equation with respect to MCS.}
\centering
\begin{tabular}{ccccc}  \hline
Models     & MAE of $\mu_{u}$ & MAE of $\sigma_{u}$ & Model evaluations & Training time (s)	\\ 
\hline
MCS & - &- & 10000&62400  \\
\PC{} & $1.06 \times 10^{-3}$ &$1.09 \times 10^{-3}$ & 0& 13800 \\
Sparse \PC{} & $4.39 \times 10^{-3}$ &$5.04 \times 10^{-3}$ & 50& 960 \\
 \hline
\end{tabular}\label{tab:Burgers UQ}
\end{table}


\subsection{Data-driven surrogate modeling: Equation of state models}
This example highlights the effectiveness of using the proposed \PC{} in a purely data-driven setting where the computational model is prohibitively expensive or when dealing with experimental data. This is often the case where we have some observational data and some understanding of the fundamental underlying physics. 

Here, we consider the data-driven and physics-informed construction of an equation of state (EOS) model. EOS models provide a mathematical relationship between a material's thermodynamic properties, such as pressure, internal energy, temperature, and density. EOS models are essential for closing the conservation equations of mass, momentum, and energy in hydrodynamic simulations, which are used across various branches
of physics, engineering, and chemistry to describe and predict the behavior of materials under different conditions. EOS models are typically constructed using semi-empirical parametric equations, which assume a physics-informed functional form with many tunable parameters that are calibrated using experimental data, first-principles simulation data, or both. Recently, there have been efforts to consider an alternate approach, in which a data-driven ML model is developed~\cite{sharma2024learning}. However, in the data-driven setting, the ML EOS model must satisfy the fundamental laws of thermodynamics necessary to make accurate predictions and be useful in hydrodynamic simulations. 

In particular, an EOS model must obey the thermodynamic stability constraints, which are derived from the second law of thermodynamics and are given as
\begin{equation}
\left(\frac{\partial^2 F}{\partial V^2}\right)_T =-\left(\frac{\partial P}{\partial V}\right)_T=\frac{1}{V \kappa_T} \geq 0 \Longleftrightarrow \kappa_T>0 \Longleftrightarrow\left(\frac{\partial P}{\partial V}\right)_T \leqslant 0, 
\label{ts_P}
\end{equation}
\begin{equation}
\left(\frac{\partial^2 S}{\partial E^2}\right)_V=-\frac{1}{T^2}\left(\frac{\partial T}{\partial E}\right)_V=-\frac{1}{T^2 c_\nu} \leqslant 0 \Longleftrightarrow  c_V>0 \Longleftrightarrow\left(\frac{\partial E}{\partial T}\right)_V \geqslant 0 ,
\label{ts_E}
\end{equation}
where $F,\ S$ are the Helmholtz free energy and entropy, respectively. $P$ and $E$ are the pressure and internal energy, respectively, which are the outputs of the EOS model. $V$ and $T$ are the volume and temperature, respectively, and are taken as model inputs. Finally, $\kappa_T$ and $c_V$ are the isothermal compressibility and specific heat, respectively, which should remain positive for all EOS predictions. 
\begin{figure}[!ht]
   \centering 
    \begin{tabular}{cc}
\includegraphics[width=0.45\textwidth]{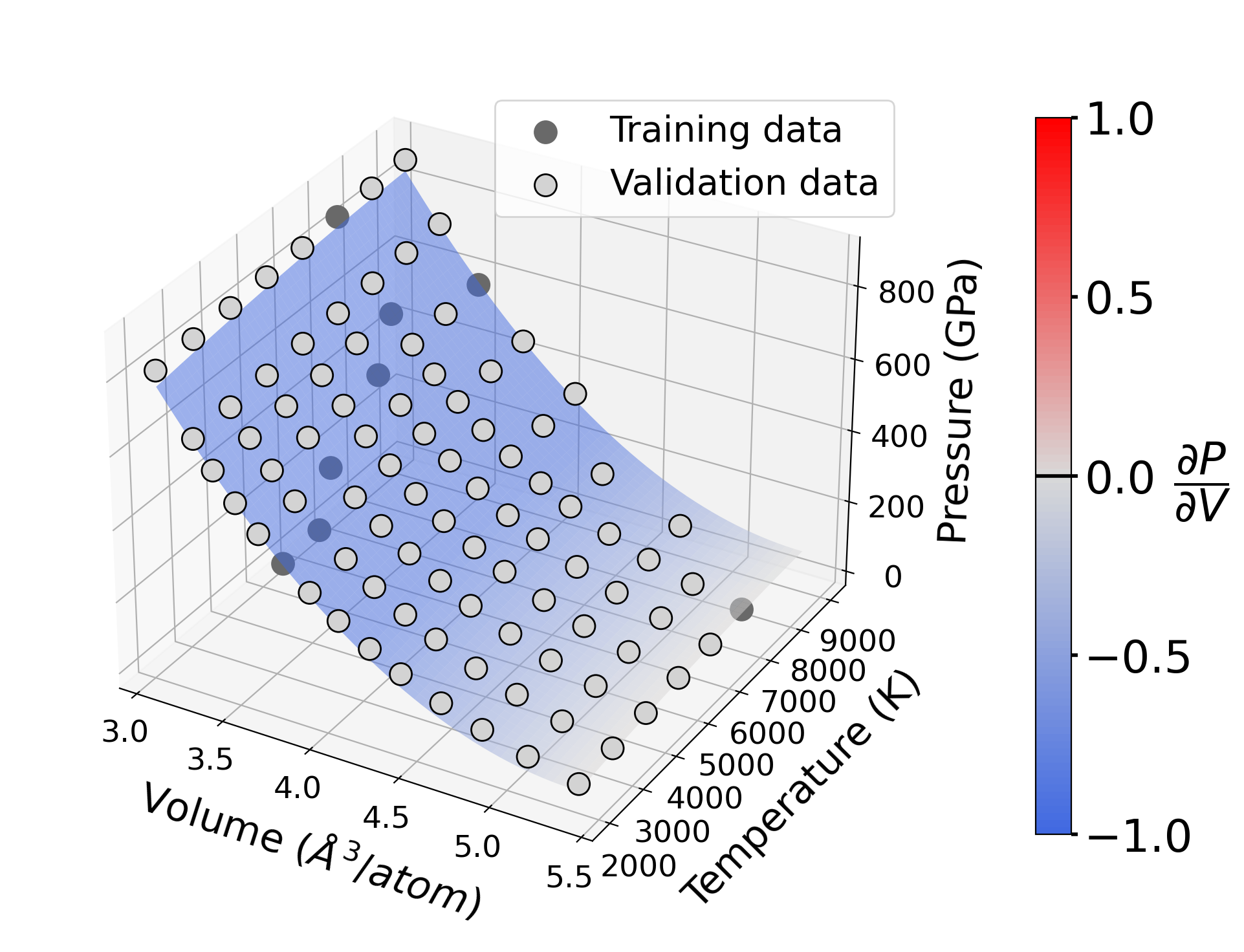} &
\includegraphics[width=0.45\textwidth]{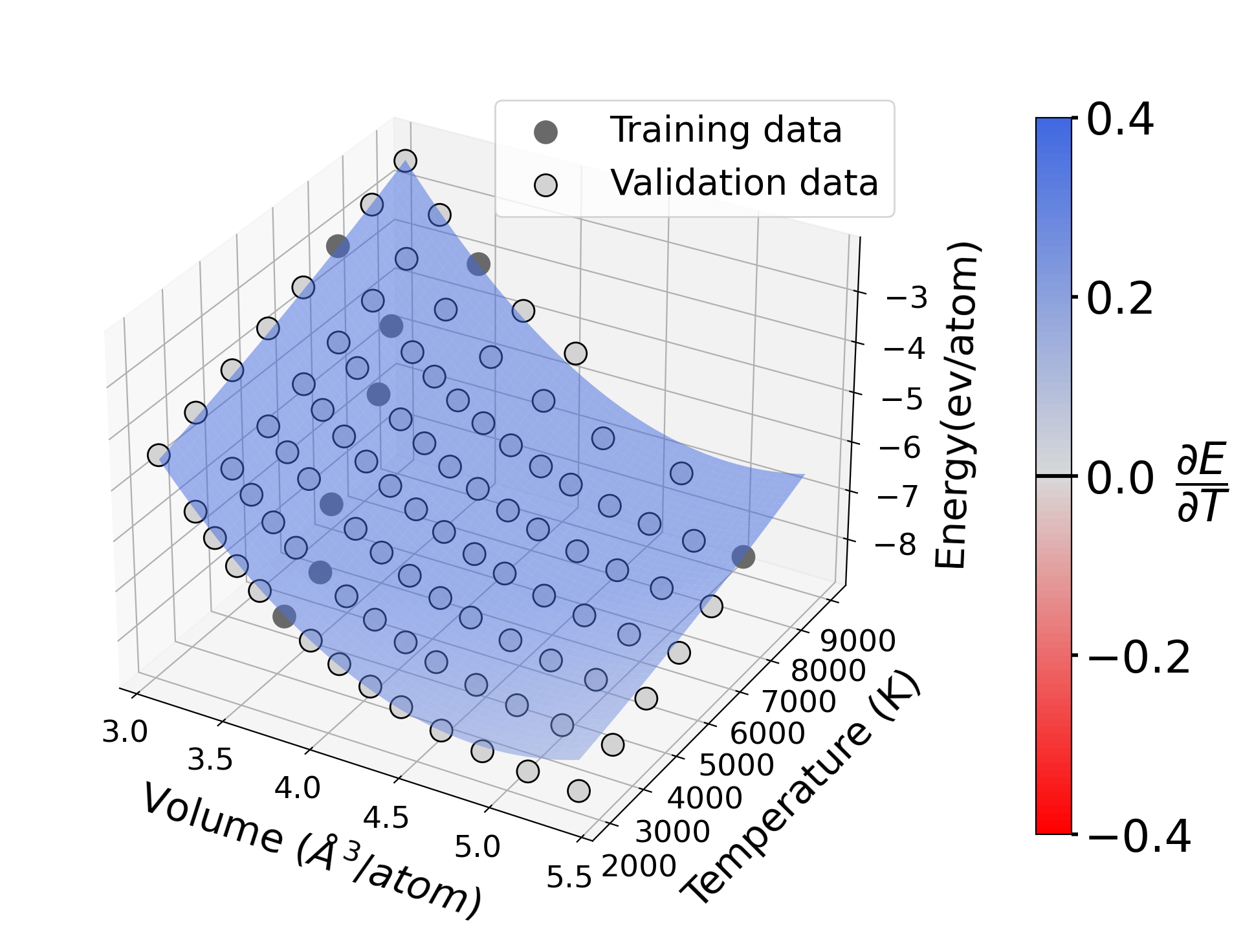} \\
(a)& (b)\\
\includegraphics[width=0.45\textwidth]{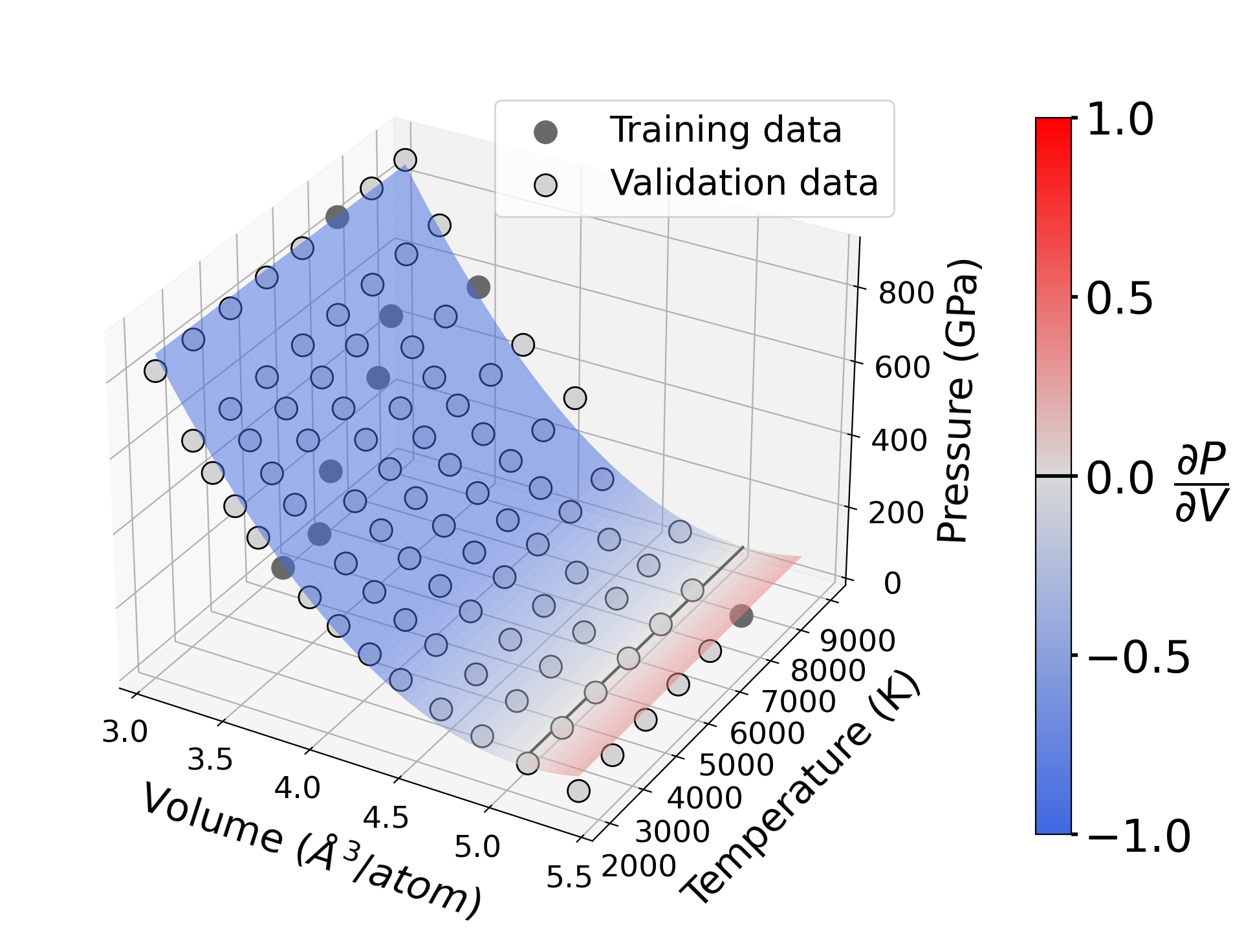} &
\includegraphics[width=0.45\textwidth]{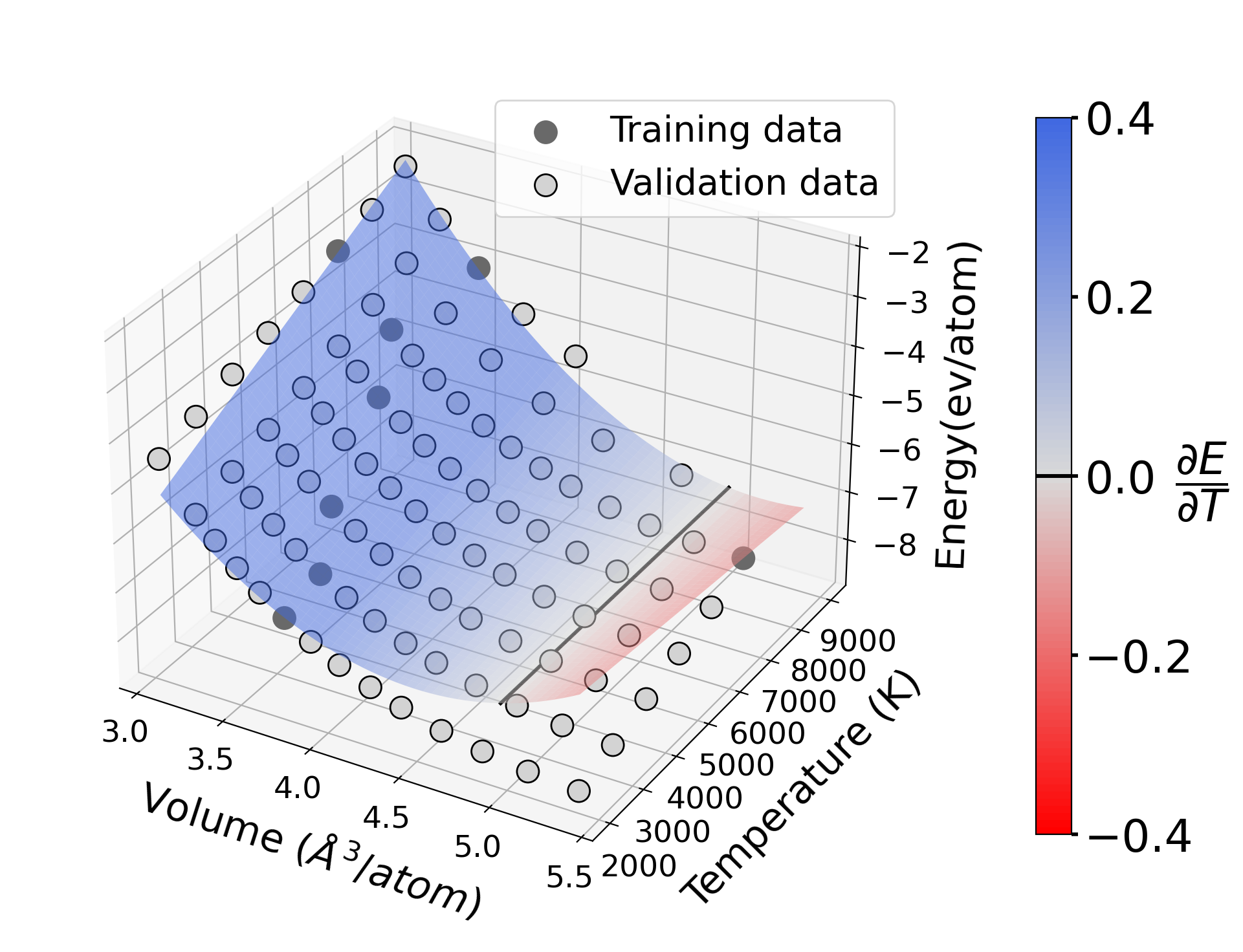} \\ 
(c)& (d)\\
\end{tabular}
\caption{Comparison of the \PC{} and standard sparse \PCE{} EOS models with regions of constraint violations shown in red. (a) Pressure \PC{} EOS model. (b) Energy \PC{} EOS model. (c) Pressure \PCE{} EOS model. (d) Energy \PCE{} EOS model. The \PC{} EOS models provide an improved fit with respect to the validation data without violating the thermodynamic constraints.}
\label{fig:EOS surrogate models}
\end{figure}

Here, we use the proposed \PC{} as a data-driven EOS model, ensuring that it satisfies the thermodynamic stability constraints given in Eqs.~\eqref{ts_P} and \eqref{ts_E}. 
The model is trained from data relating $(P,E)$ and $(V,T)$ generated using Density Functional Theory Molecular Dynamics (DFT-MD) simulations for diamond from Benedict et al.~\cite{benedict2014multiphase}. 
We have 96 data points, which we split into training and validation sets of 8 and 88 points, respectively. Figures \ref{fig:EOS surrogate models} (a) and (b) show the \PC{} models for pressure and energy as a function of state variables $(V, T)$ trained from 8 data points, with color bars representing the constraint violations. We enforce the constraints at 600 virtual collocation points in the input domain and use $p=2$. We similarly trained using a standard sparse \PCE{} model in Figures \ref{fig:EOS surrogate models} (c) and (d). It can be clearly observed that the \PC{} provides a better fit for both $P$ and $E$ with respect to the validation set compared to the sparse \PCE{}. Also, it is evident from the figures that the proposed \PC{} surrogate model satisfies the constraints at every point in the input domain. However, the standard sparse \PCE{} surrogate model violates both constraints at many points, which suggests physically unrealistic predictions, making it impractical to use in hydrodynamic simulations. 

\begin{figure}[!ht]
   \centering 
    \begin{tabular}{cc}
\includegraphics[width=0.4\textwidth]{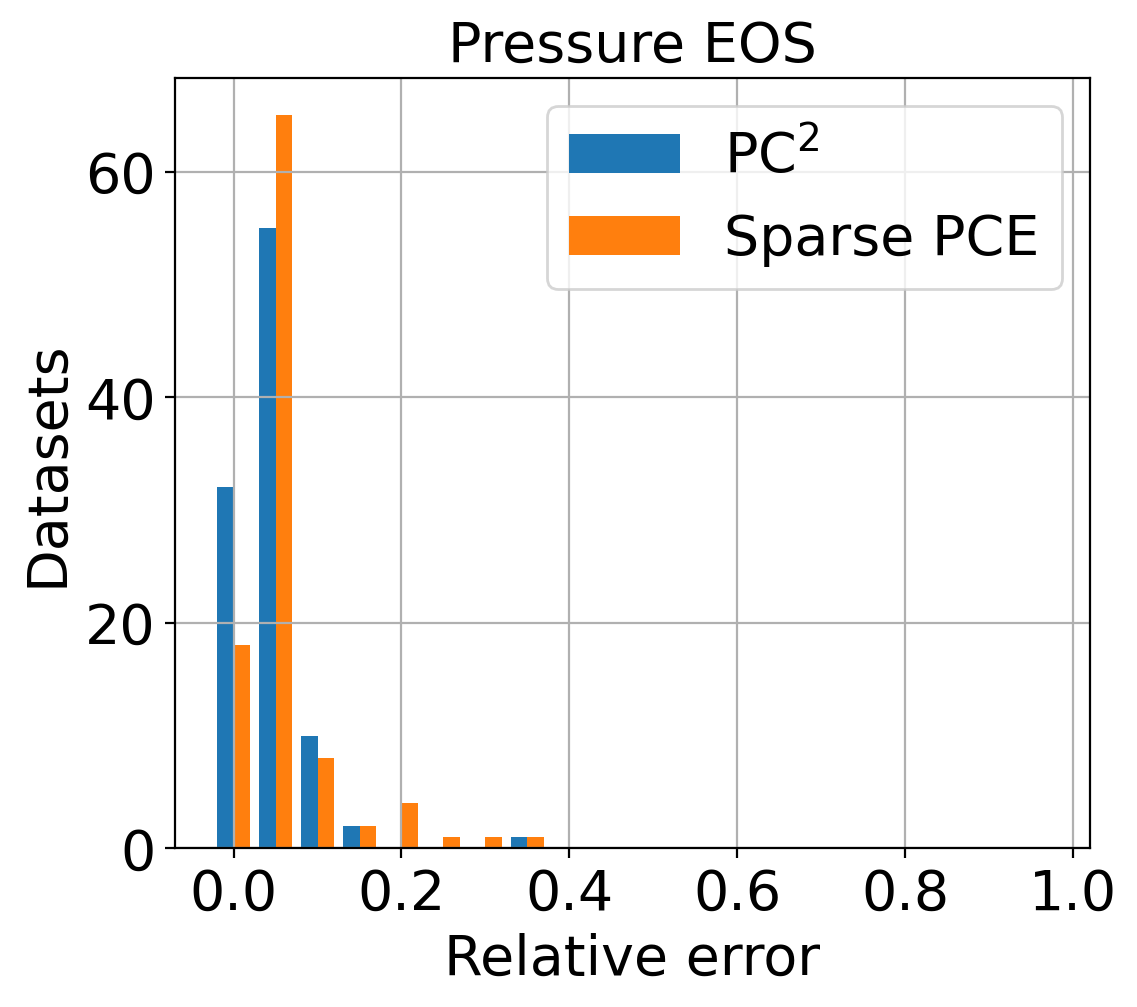} &
\includegraphics[width=0.4\textwidth]{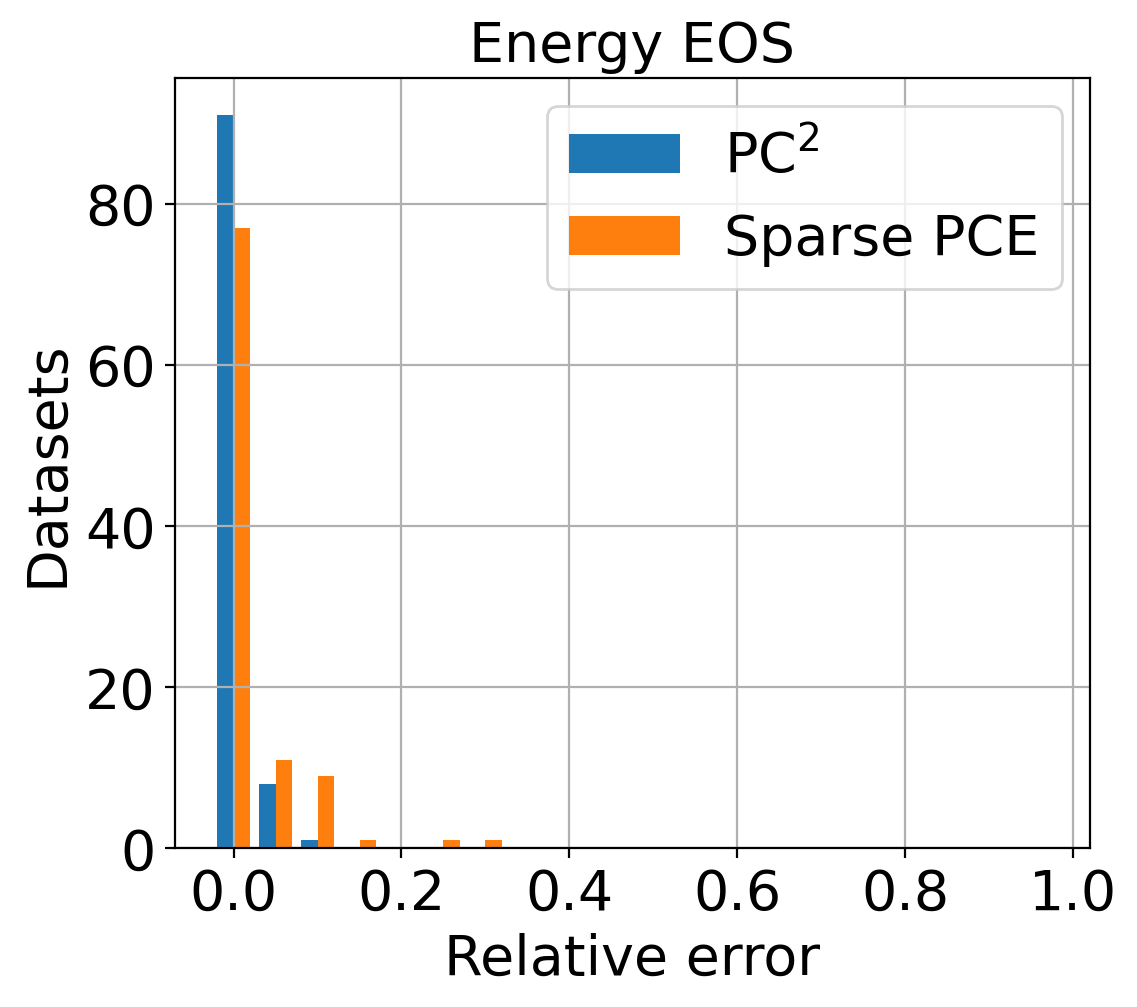} \\
\multicolumn{2}{c}{(a)}  \\
\includegraphics[width=0.4\textwidth]{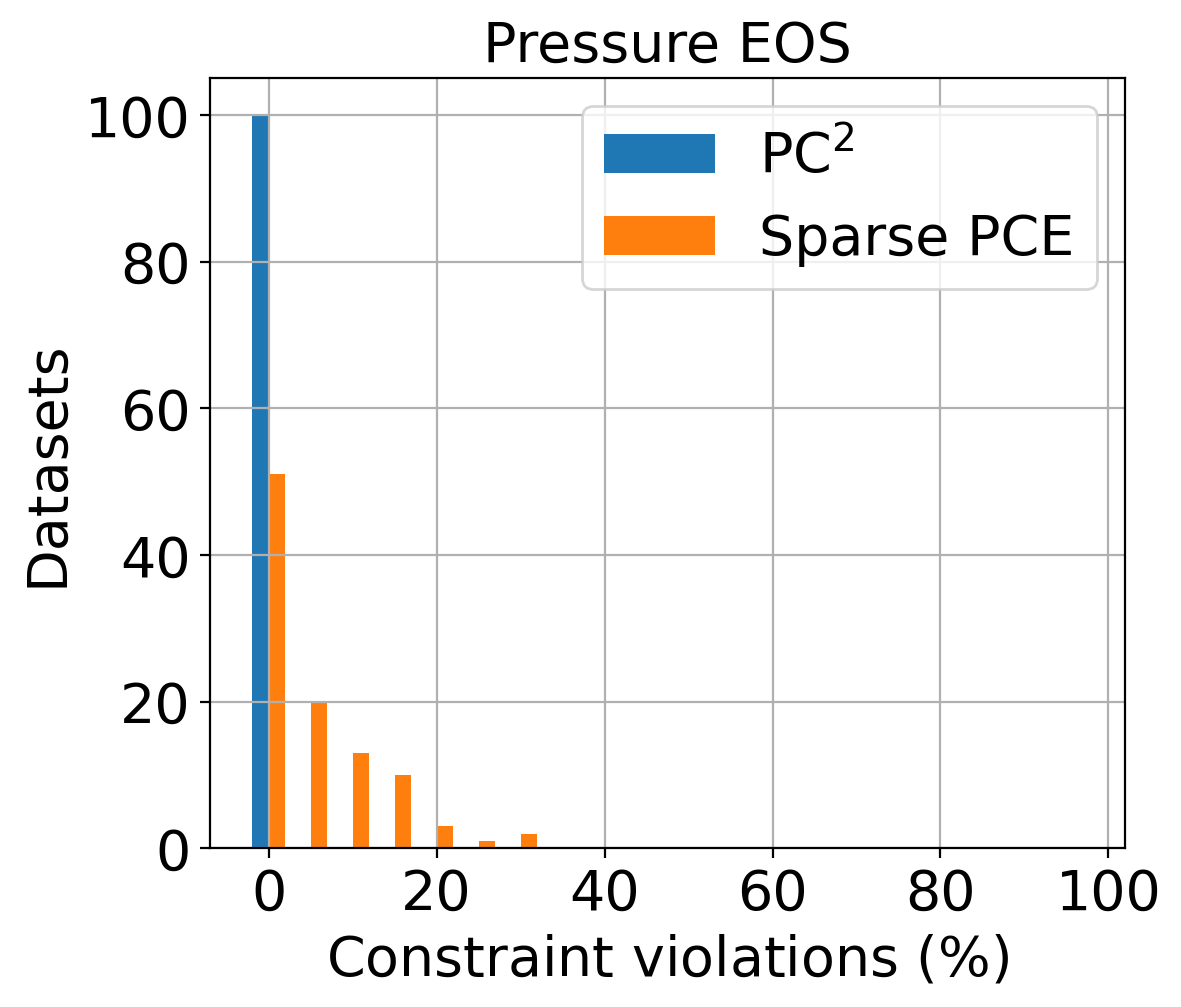} &
\includegraphics[width=0.4\textwidth]{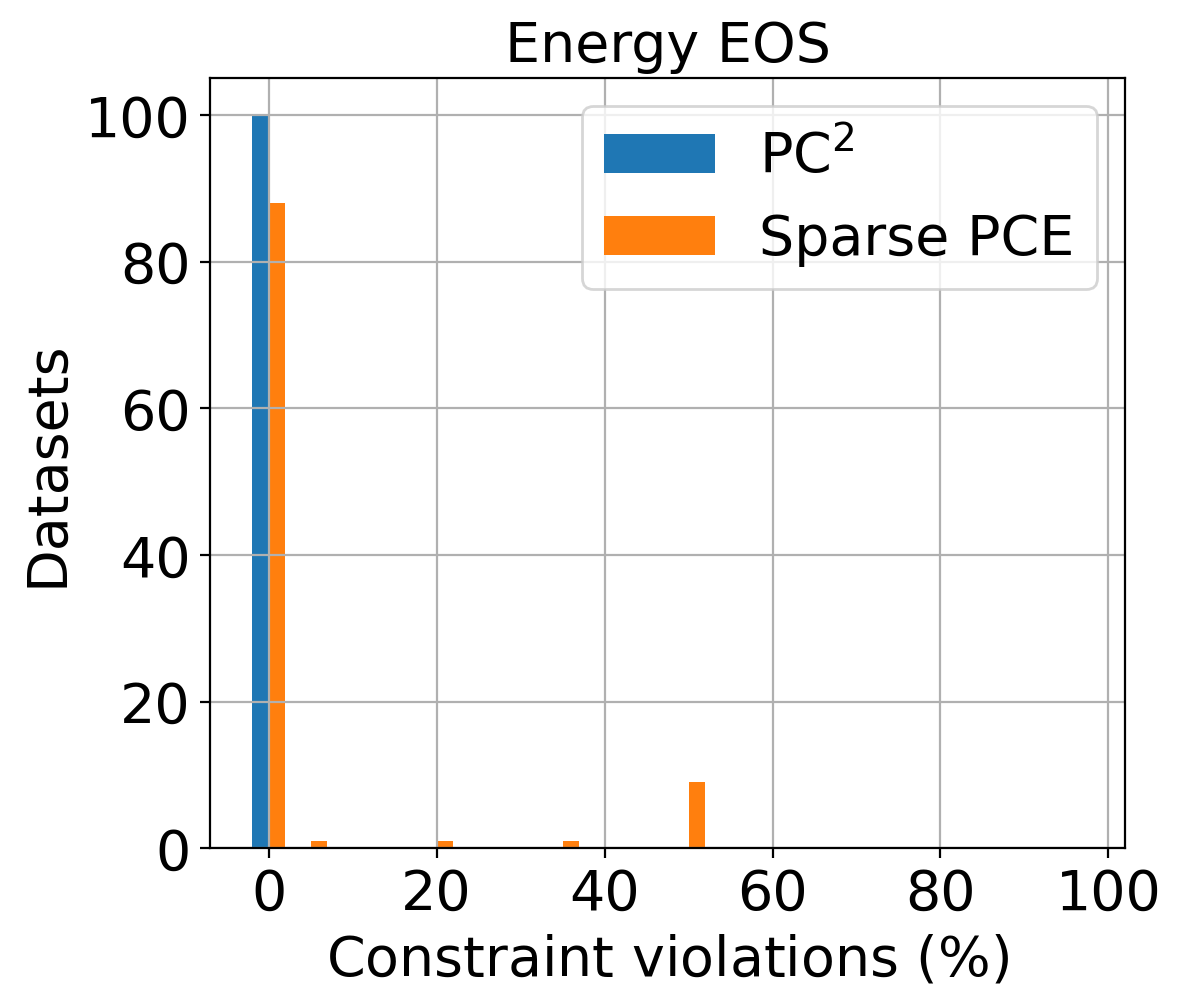} \\
\multicolumn{2}{c}{(b)} 
\end{tabular}
\caption{Histograms of relative $\pazocal{L}^2$ error and percent constraint violations for 100 \PC{} and sparse \PCE{} models trained from datasets with 8 randomly selected training points from the given 96 DFT-MD data. (a) $\pazocal{L}^2$ error for pressure EOS and energy EOS, showing more datasets with low relative error for \PC{} compared to sparse \PCE{}. (b) Constraint violations (\%) for the pressure EOS and energy EOS, showing all datasets satisfy the constraints for \PC{} compared to many datasets violating constraints for sparse \PCE{}. $\pazocal{L}^2$ error is calculated for each training dataset with respect to the complementary validation dataset. Constraint violations are shown as the percentage of the test points violating the constraints.}
\label{fig:EOS histograms}
\end{figure}

\begin{figure}[!ht]
   \centering 
    \begin{tabular}{cc}
\includegraphics[width=0.45\textwidth]{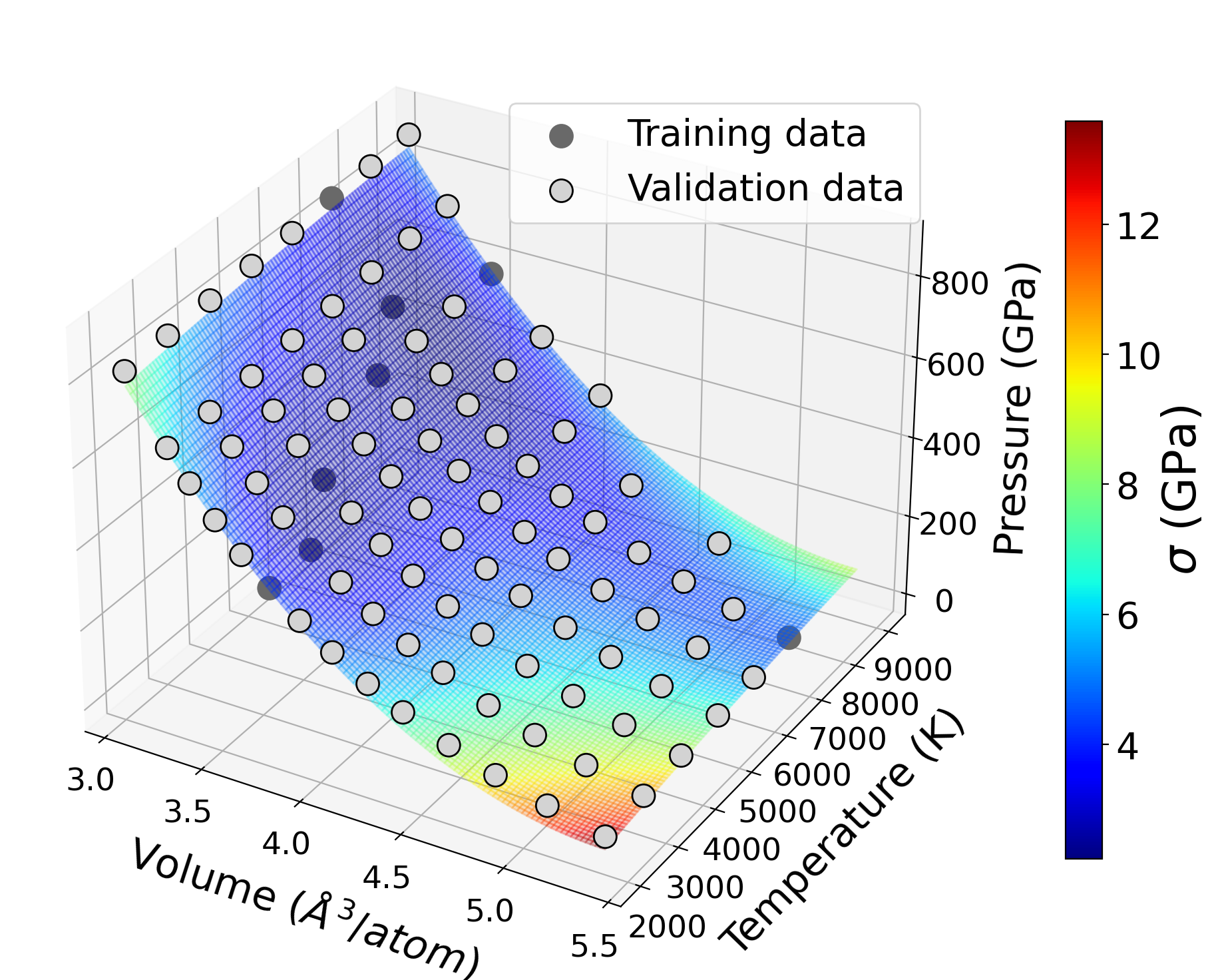} &
\includegraphics[width=0.45\textwidth]{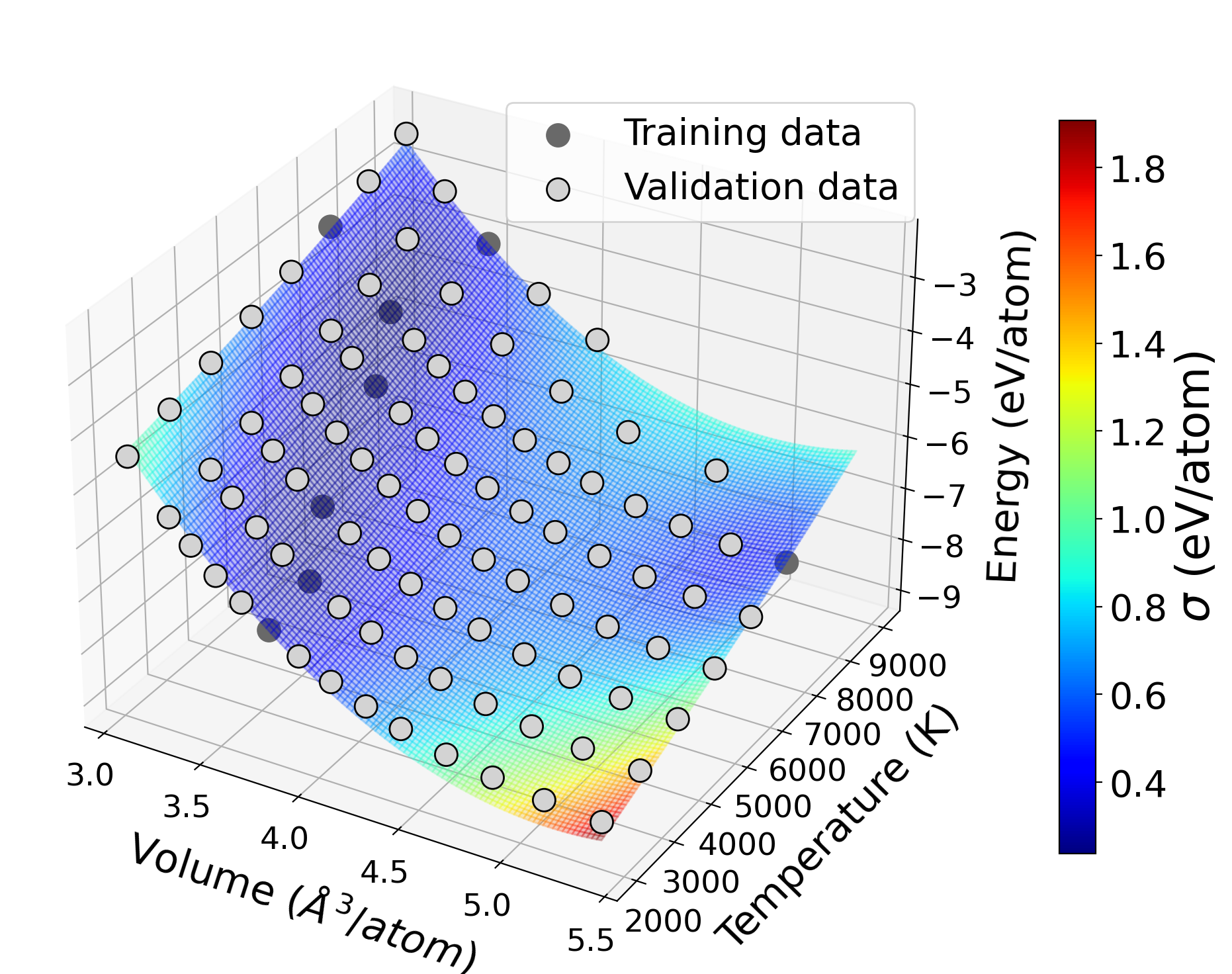} \\
(a)& (b)\\
\includegraphics[width=0.45\textwidth]{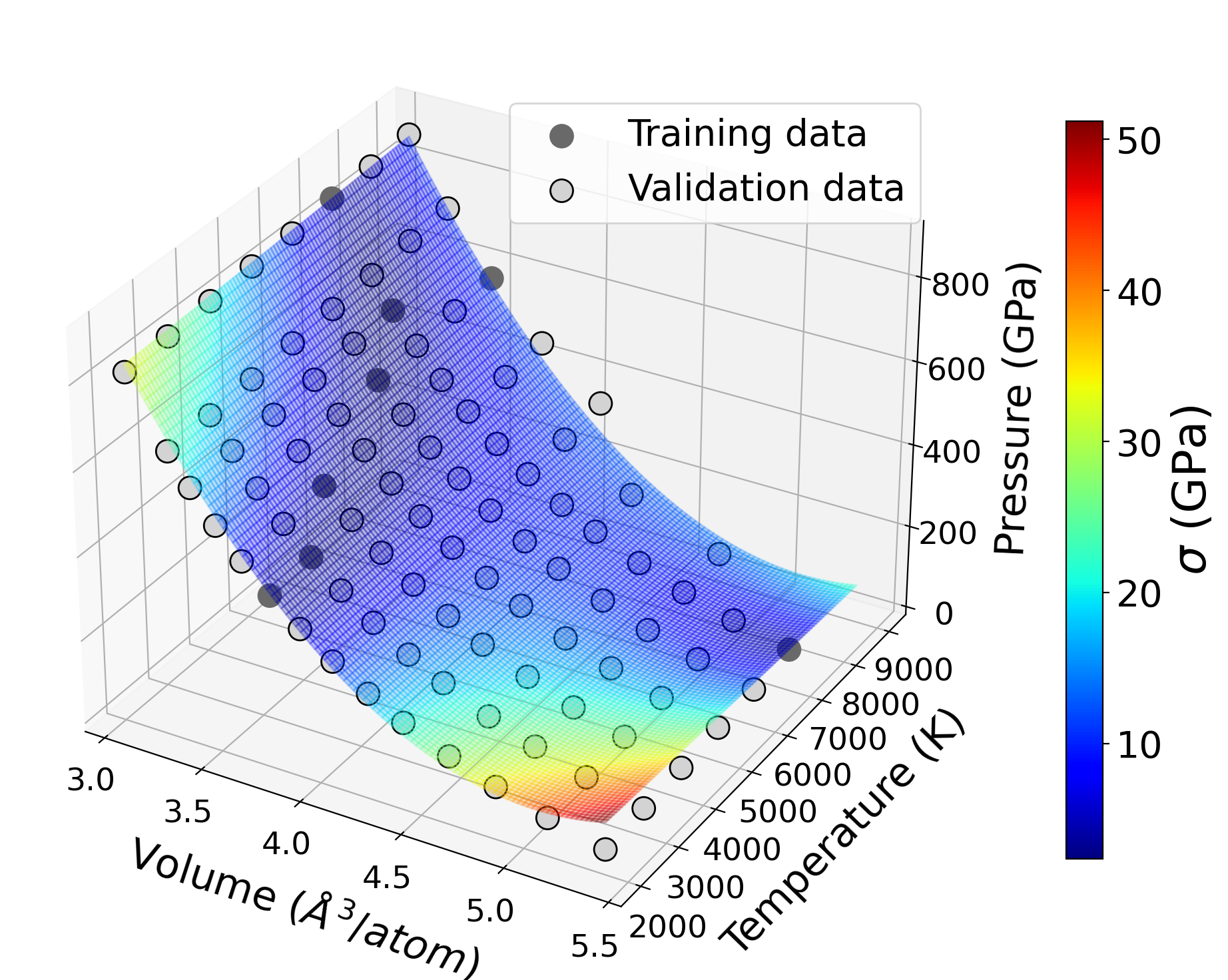} &
\includegraphics[width=0.45\textwidth]{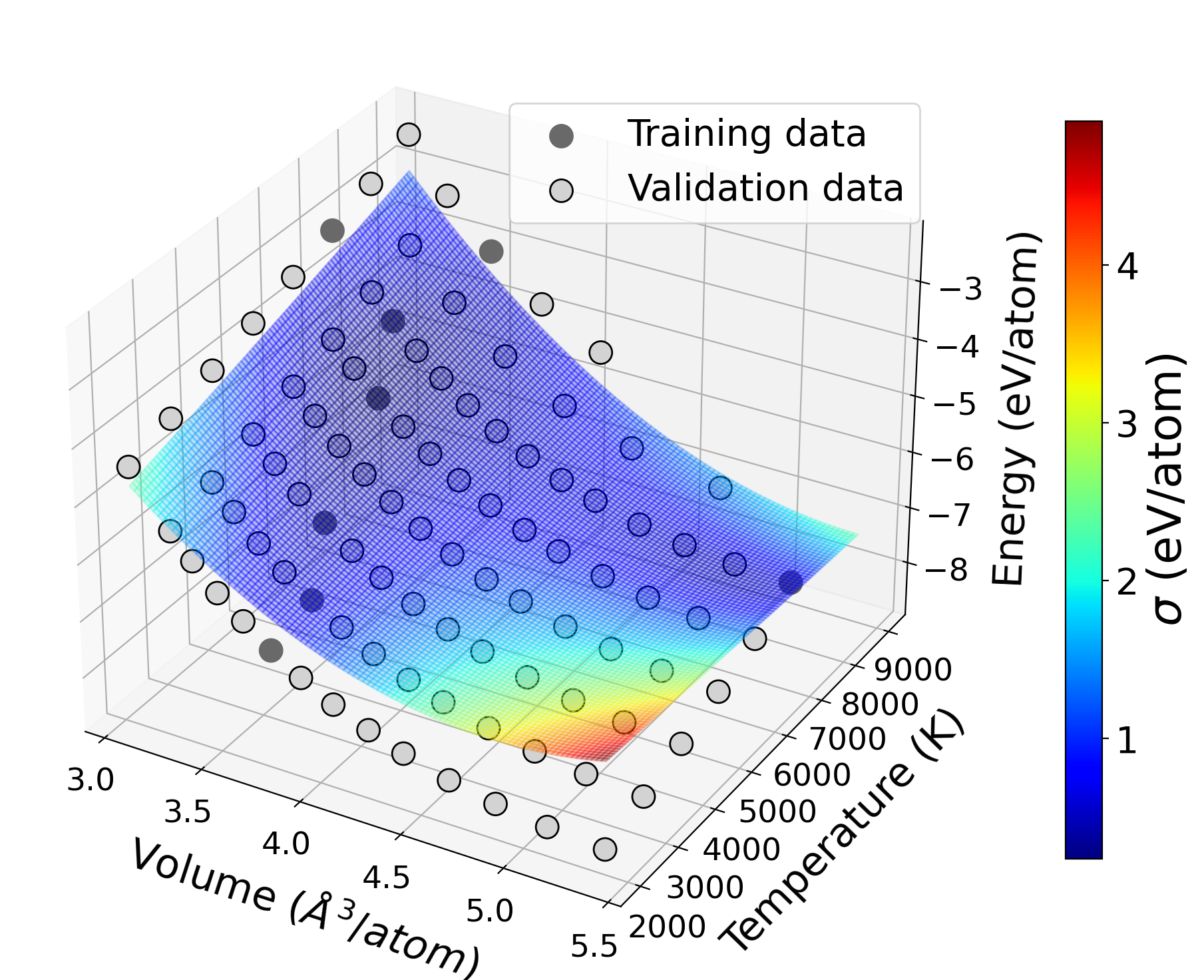} \\ 
(c)& (d)\\
\end{tabular}
\caption{Comparison of the \PC{} and sparse \PCE{} EOS models trained with Gaussian noise added to the output observations of the training data. (a) Mean pressure \PC{} EOS. (b) Mean energy \PC{} EOS. (c) Mean pressure \PCE{} EOS. (d) Mean energy \PCE{} EOS. Colors denote the standard deviation of the model at each point. The mean \PC{} models provide an improved fit with respect to the validation data with significantly lower standard deviation compared to sparse \PCE{}, indicating enhanced robustness in handling noise in the training data.}
\label{fig:EOS_noise}
\end{figure}

To demonstrate the robustness of the \PC{} model, we created 100 datasets by randomly selecting 8 training points from the 96 DFT-MD data points and using the complementary data as validation sets. For each dataset, we trained the proposed \PC{} and standard sparse \PCE{} models. For each trained model, we report the relative $\pazocal{L}^2$ error with respect to the respective validation set and the percentage of constraint violations in the histograms in Figure \ref{fig:EOS histograms}. Figure \ref{fig:EOS histograms} (a) shows that the histograms are weighted more toward the lower relative error for \PC{}, which suggests an improved fit compared to sparse \PCE{}. Meanwhile, Figure \ref{fig:EOS histograms} (b) shows that the \PC{} model does not violate the constraints for any of the datasets, whereas the standard sparse \PCE{} model violates the constraints violations at many points (sometimes exceeding 50\% of the points). These plots clearly indicate superior performance of the \PC{} model in a data-driven setting compared to the standard sparse \PCE{}. 

Finally, we study the behavior of the proposed \PC{} model in handling noise in the training data compared to sparse \PCE{} by adding Gaussian noise having zero mean and standard deviation 5 GPa (for $P$) / 0.5 eV/atom (for $E$) to each observation in the training dataset. The model is then retrained using the perturbed outputs, and the predictions are ensembled to compute statistics \cite{torre2019data}. 
Figure~\ref{fig:EOS_noise} shows the mean \PC{} and standard sparse \PCE{} EOS models with standard deviation shown in color for pressure and internal energy against volume and temperature. The mean EOS using \PC{} provides a better fit with respect to the validation data points for each output compared to sparse \PCE{}.  
Moreover, the standard deviation of the \PC{} EOS is significantly less than the sparse \PCE{}, which suggests enhanced robustness in handling noise in the training data. As expected, the standard deviation is higher in regions where the training data is not present. The superior performance of \PC{} in this data-driven setting is attributed to the incorporation of physical constraints, which enriches the limited training data and yields more realistic predictions.

\subsection{Uncertainty Quantification: Stochastic Euler Bernoulli beam}

In the final example, we employ the \PC{} surrogate model in a standard \UQ{} setting where \PCE{} is typically employed. The objective of this example is to compare the performance of the proposed \PC{} method with the standard sparse \PCE{} in \UQ{} of physical systems with increasing stochastic dimension.
Here, we apply \PC{} to perform UQ for an Euler Bernoulli (EB) beam with Young's modulus ($E(\mathbf{x}, \theta)$) modeled as a 1D Gaussian random field discretized with the Karhunen-Loève (KL) expansion \cite{ghanem2003stochastic} and having exponential covariance kernel; $C(x_{1},x_{2})=\sigma^{2}\exp(-|x_1-x_2|/l_{c})$, where $\sigma$ is the standard deviation, $x_1$ and $x_2$ are coordinates along the length of the beam and $l_c$ is the correlation length of the random field. 

\begin{figure}[!ht]
   \centering 
    \begin{tabular}{cc}
\includegraphics[width=0.8\textwidth]{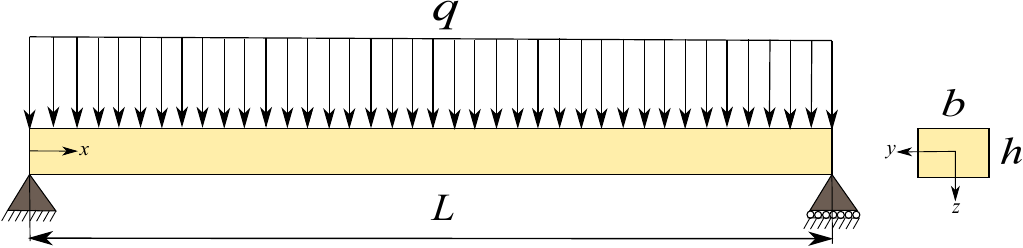} 
\end{tabular}
\caption{Simply supported Euler Bernoulli beam.}
\label{fig:EB beam diagram}
\end{figure}

A simply supported Euler Bernoulli beam of length $L$ with uniformly distributed load $q$, is shown in Figure~\ref{fig:EB beam diagram}. The bending behavior of the EB beam is governed by the following equation:
\begin{equation}
\frac{\mathrm{d}^2}{\mathrm{~d} x^2}\left(E I \frac{\mathrm{d}^2 w}{\mathrm{~d} x^2}\right)=q,
\end{equation}
where $w$ denotes the lateral deflection, and $I$ is the moment of inertia of the beam's cross-section.

\begin{figure}[!ht]
    \centering
    \includegraphics[width=0.45\textwidth]{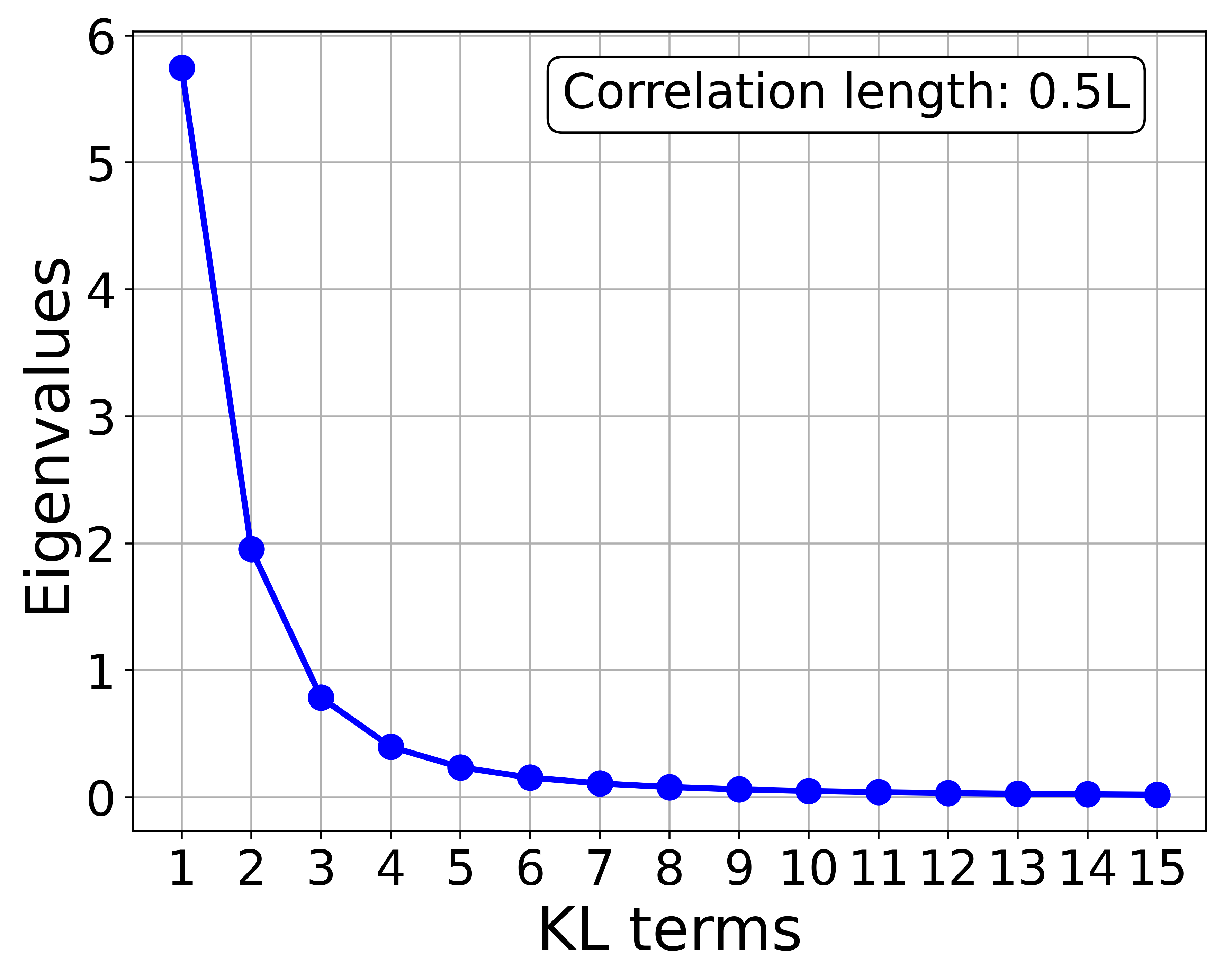}
    \caption{Eigenvalue decay with respect to the number of KL terms indicating 7 KL terms are sufficient to capture the stochasticity of the random field for $l_c=0.5L$.}
    \label{fig:eigen_value decay}
\end{figure}

The KL expansion discretization for the $E(\mathbf{x}, \theta)$ random field is given as
\begin{equation}
    E(\mathbf{x}, \theta)= {\overline {E}}+\sum_{i=1}^{r} \sqrt{\lambda_{i}} \phi_{i}(\mathbf{x}) \xi_{i}(\theta),
    \label{eq:KL_expansion}
\end{equation}
where ${\overline {E}}$ is the mean of the random field, $\left(\lambda_i, \phi_i\right)$ are the eigenpairs, $\xi_i(\theta)$ are independent standard Gaussian random variables, and $r$ represents the total number of terms in the truncated expansion. 
For our computations, we considered $L=10\ m$, $q=-5\ kN/m$, $I=10^{-4}\ m^4$, and ${\overline {E}}=80\ GPa$. 
Figure \ref{fig:eigen_value decay} shows the eigenvalue decay with respect to the number of KL terms for a correlation length $l_c=0.5L$. As can be observed, $r=7$ KL terms are sufficient to capture the stochastic behavior of the underlying random field.

\begin{figure}[!ht]%
 \centering
  \captionsetup[subfloat]{labelfont=normalsize,textfont=normalsize}
 \subfloat[]{\includegraphics[width=0.45\textwidth]{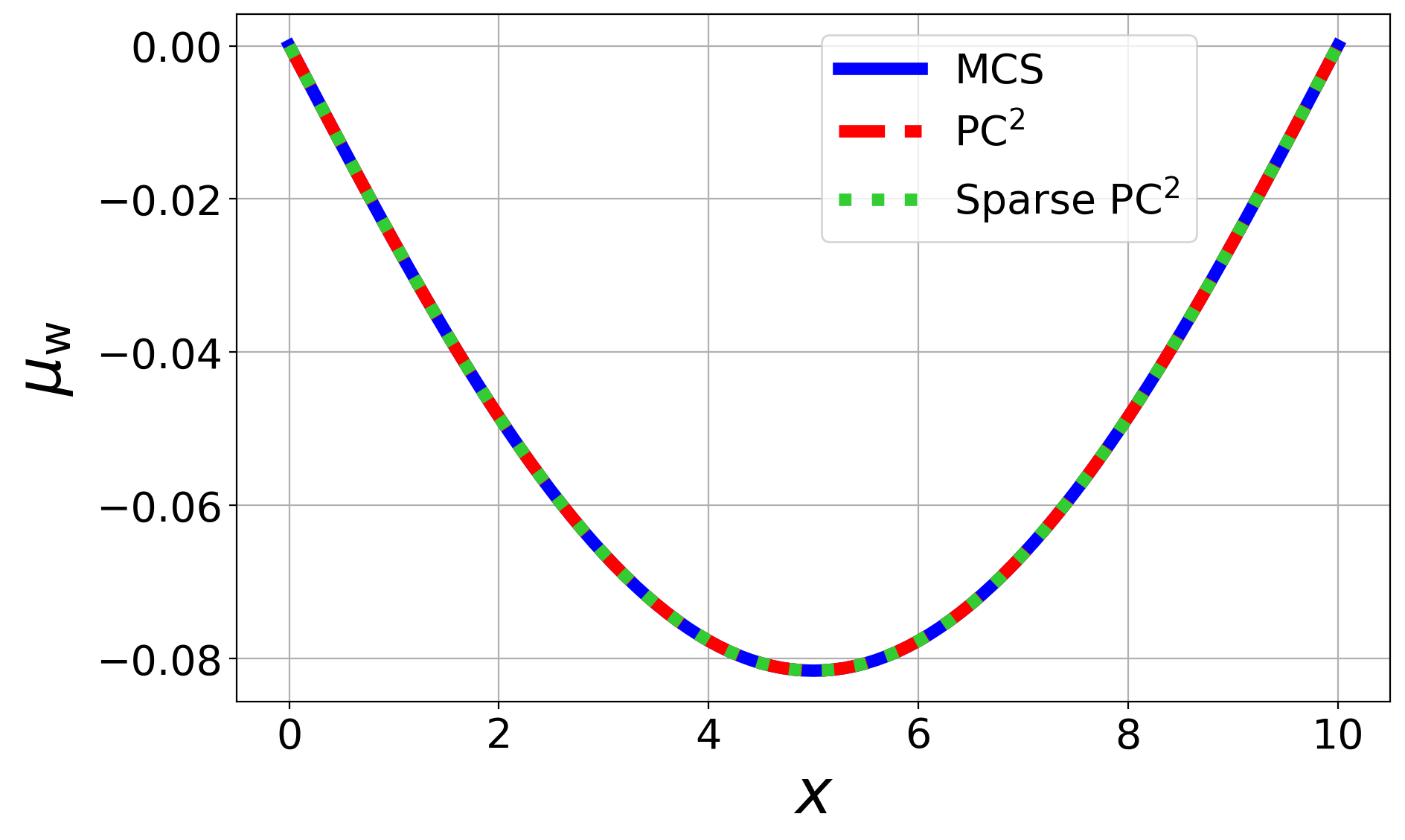}}\label{fig:a}
 \subfloat[]{\includegraphics[width=0.45\textwidth]{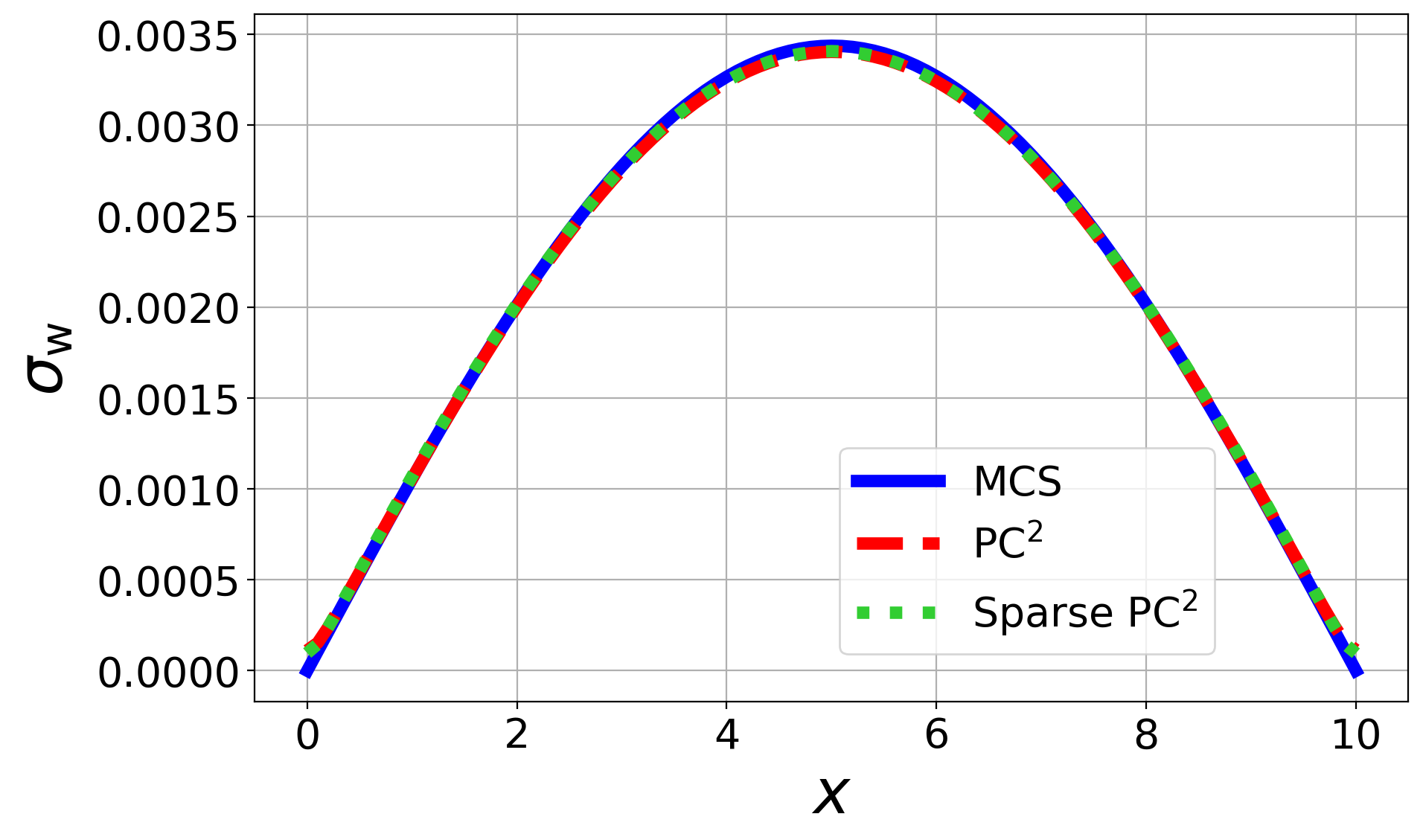}}\label{fig:b}\\
 \subfloat[]{\includegraphics[width=0.45\textwidth]{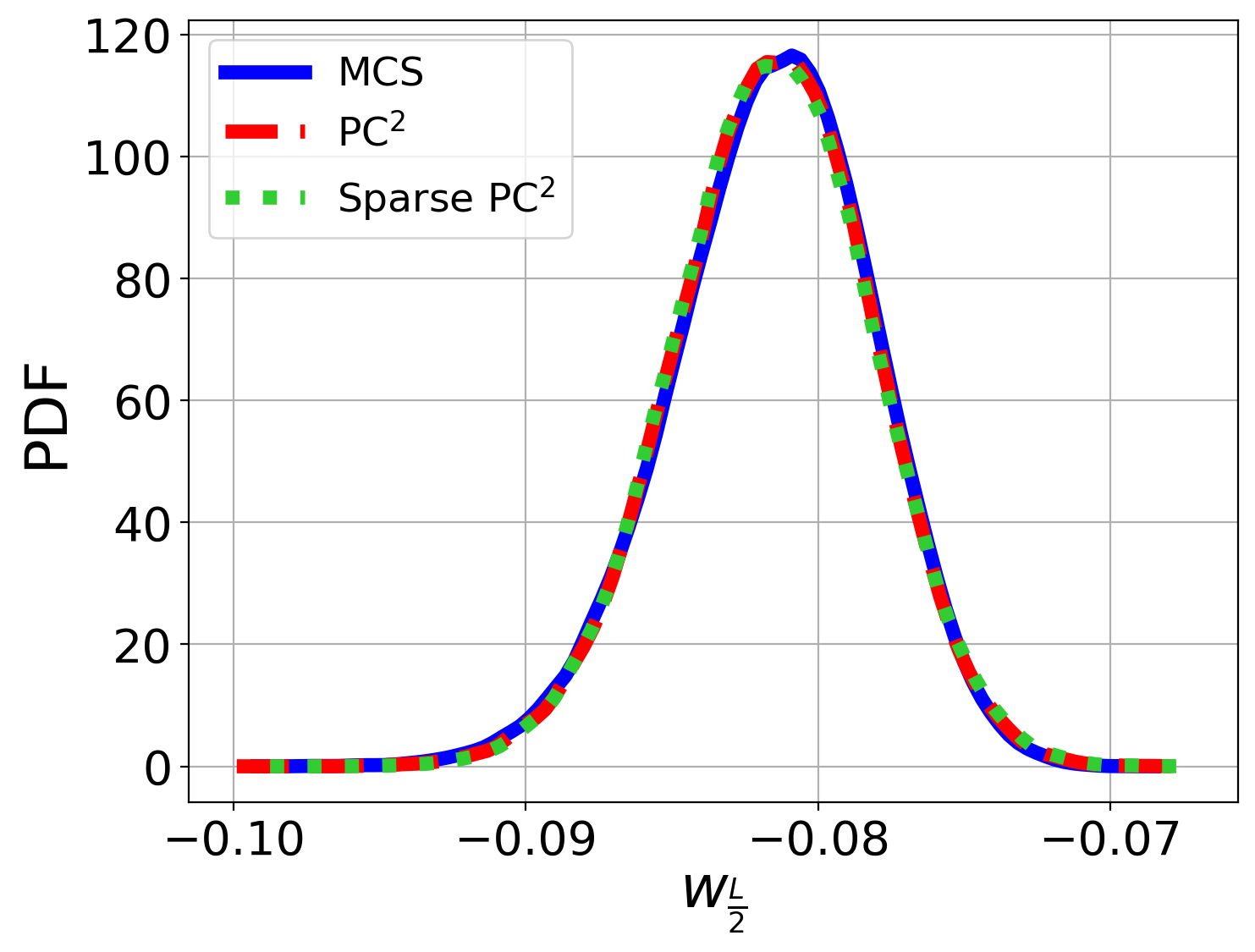}\label{fig:c}}%
 \caption{Euler-Bernouilli Beam: (a) Plot of the mean deflection along the length of the beam estimated by  MCS, \PC{} and sparse \PC{}. (b) Plot of the standard deviation of the deflection along the length of the beam estimated by  MCS, \PC{} and sparse \PC{}. (c) Plot of the probability density function of the midpoint deflection estimated by  MCS, \PC{} and sparse \PC{}. All the plots indicate excellent convergence of \PC{} and sparse \PC{} in estimating the response statistics in comparison to MCS. We used zero model evaluations for \PC{}, 100 model evaluations for sparse \PC{}, and 100,000 model evaluations for MCS with $\text{COV} = 5\%$.}%
 \label{fig:EB_beam_statistics}%
\end{figure}

For the \PC{} model, we take $p=5$ and the number of virtual collocation points as  $n_{v}=20000,\ n_{v}^{\text{\scriptsize BC}}=5000$ for both the physical and stochastic dimensions.
Figures \ref{fig:EB_beam_statistics} (a) and (b) shows the the mean and standard deviation of the deflection $w(x)$ computed using \PC{}, sparse \PC{} and MCS. To train the surrogate models, we used zero model evaluation for \PC{} and 100 model evaluations for the sparse \PC{}, compared to 100,000 evaluations for MCS. The plots show that the proposed \PC{} provides an excellent statistical approximation of both the mean and standard deviation across $x$ compared to MCS. Figure \ref{fig:EB_beam_statistics} (c) shows the probability density function (pdf) of the midpoint deflection estimated using \PC{}, sparse \PC{}, and MCS, demonstrating excellent convergence and indicating superior performance in capturing the stochasticity of the response. Also, it can be observed from the figures that the numerical accuracy is similar for both the \PC{} and sparse \PC{}; however, the sparse \PC{} took only 25 seconds compared to 1440 seconds for \PC{} to train, which shows significant computational savings for sparse \PC{}. Hence, it is recommended to use sparse \PC{} when dealing with high-dimensional problems.

\begin{figure}[!ht]
    \centering
    \includegraphics[width=0.9\textwidth]{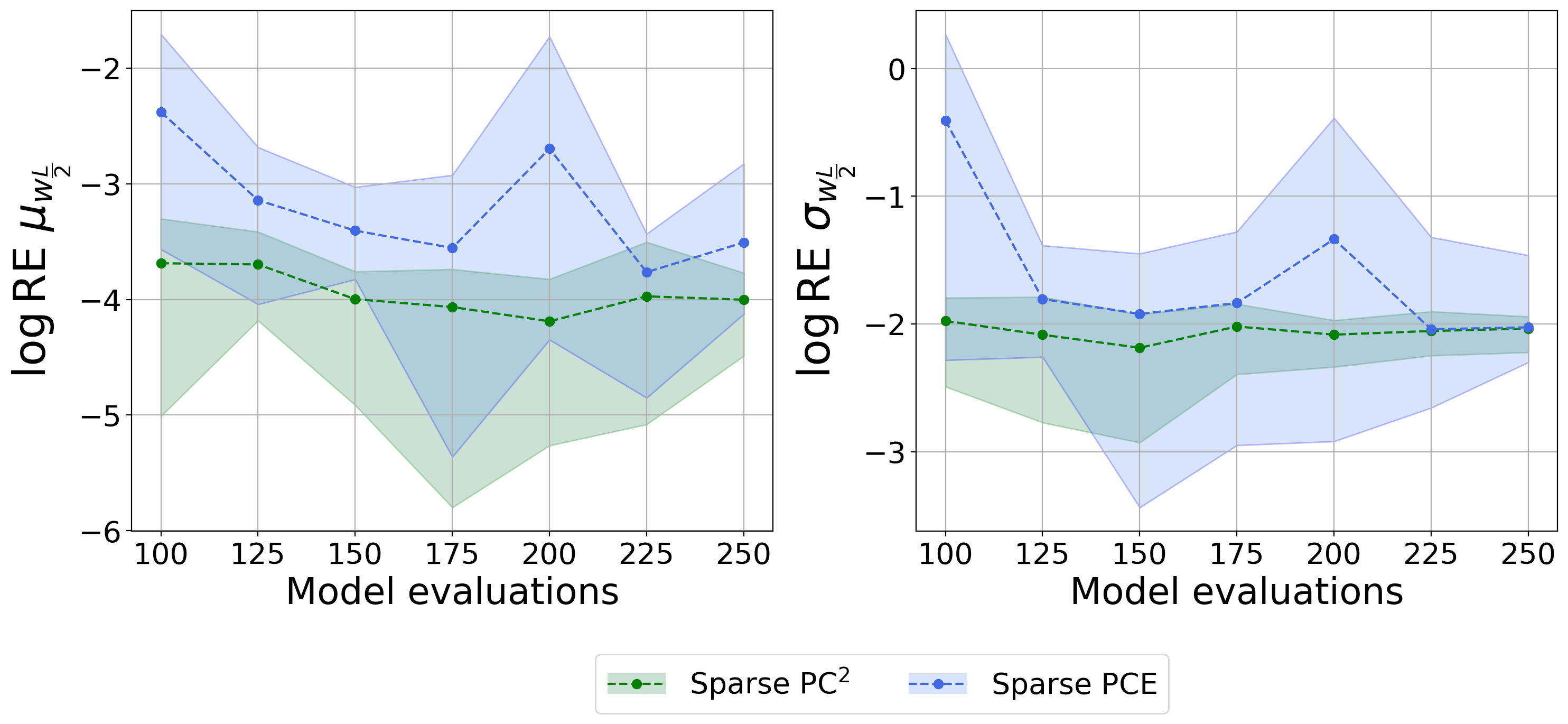}
    \caption{Euler-Bernoulli Beam: Comparison of the relative error of the mean and standard deviation of midpoint deflection estimated using sparse \PC{} and sparse \PCE{} with respect to MCS for increasing number of model evaluations. The dotted line indicates the average relative error, and the shaded area represents the minimum and maximum error across 10 repetitions for each model evaluation. ($\text{COV}=10\%$)}
    \label{fig:beam_conv}
\end{figure}

Next, to compare the performance of sparse \PC{} with sparse \PCE{}, we computed the relative error in mean and standard deviation of the mid-point deflection for sparse \PC{} and sparse \PCE{} with respect to MCS. To check the robustness of performance, we retrained both surrogate models 10 times
for each number of model evaluations 
with ED based on LHS sampling. The results are shown in Figure~\ref{fig:beam_conv}, with dotted lines representing the average relative error and the shaded area representing the range from the minimum to the maximum relative error observed in the 10 repetitions. It can be observed from the figure that the sparse \PC{} provides improved performance in terms of the average relative error for both mean and standard deviation. Moreover, from the minimum and maximum relative errors, it can be inferred that the sparse \PC{} is less sensitive to the sampling of the ED, which is expected since the \PC{} is enriched by physical constraints and provides a consistent performance regardless of the number of model evaluations. On the other hand, the sparse \PCE{} is sensitive to the sampled points in ED, especially for a smaller number of model evaluations, and the performance improves with the increasing number of model evaluations. Also, the maximum relative error for sparse \PC{} is much less than the sparse \PCE{}, suggesting reliable estimations of the moments for sparse \PC{}. This demonstrates the superior performance of the proposed \PC{} surrogate method in the reliable and robust uncertainty assessment of stochastic systems with fewer model evaluations. 

Finally, to test the computational performance of the proposed sparse \PC{}, we further increased the stochastic dimension by decreasing the correlation length, $l_c$, and correspondingly increasing the number of terms in the KL expansion. We consider 10, 15, and 20 KL terms, with $p=5$ and one physical variable, which corresponds to the total basis with 4368, 20349, and 65780 polynomials, respectively. We considered 300 model evaluations for training in each case and reported the computational time, mean, and standard deviation of the midpoint deflection estimated by sparse \PC{} and MCS in Table \ref{tab:beam_comp_sparse}. It can be observed from the table that the sparse \PC{} provides excellent computational performance for problems with high stochastic dimension, taking at most a few minutes to achieve a low target error, as evident from the accuracy of estimated means and standard deviations compared to MCS. This improvement is attributed to the simplified sparse implementation and the incorporation of physical constraints. Despite a very high number of polynomial basis functions in the full expansion, only a few hundred contribute most to the output response, which are effectively identified by the sparse \PC{} through LAR. This indicates that the proposed method is suitable for efficient and reliable \UQ{} of stochastic systems with high stochastic dimension. However, it is worth mentioning that the performance of the sparse \PC{} depends on the effectiveness of LAR to provide the most influential polynomial basis, which generally improves with increasing number of model evaluations. For cases with insufficient model evaluations, the sparse \PC{} may require more polynomial basis functions to achieve a given target error, resulting in an increased computational cost.   
\begin{table}[!ht]
\caption{Euler-Bernoulli Beam: Computational performance and accuracy of the sparse \PC{} for increasing stochastic dimension ($\text{COV}=10\%$) with 300 model evaluations used for training with $p=5$.} \label{tab:beam_comp_sparse}
\vspace{0.2cm}
\centering
\begin{tabular}{cccccccc}  \hline
KL terms &$l_{c}$  & Training  & \multicolumn{2}{c}{Sparse \PC{}} &	\multicolumn{2}{c}{MCS}
\\ \cline{4-7}  
& & time (s)& \multicolumn{1}{c}{$\mu_{w_{0.5L}}$} &\multicolumn{1}{c}{$\sigma_{w_{0.5L}}$} &\multicolumn{1}{c}{$\mu_{w_{0.5L}}$} &\multicolumn{1}{c}{$\sigma_{w_{0.5L}}$}\\
\hline

10 &$0.3L$   & 26 & -0.08213708 & 0.00627731 & -0.08215744 & 0.00638659 \\
15 &$0.2L$  & 95 & -0.08211130 & 0.00564987 & -0.08216044 & 0.00574360 \\
20 &$0.1L$ & 382 & -0.08202917 & 0.00447343 & -0.08212911 & 0.00411479 \\
 \hline
\end{tabular}
\end{table}

\section{Conclusion}

In this work, we have proposed a novel physics-constrained polynomial chaos (\PC{}) surrogate model that integrates physical knowledge into the polynomial chaos framework. The incorporation of physical constraints in the training process effectively reduces (and sometimes eliminates) the need for expensive model evaluations and ensures physically realistic predictions. The proposed method is capable of addressing problems related to scientific machine learning and uncertainty quantification, which we have demonstrated through various numerical examples. In the scientific machine learning domain, the proposed method is well-suited for problems with some data and a partial understanding of the underlying physics and features a built-in uncertainty quantification capability. In uncertainty quantification domains, the incorporation of physics constraints enriches the experimental design, enhancing the accuracy of uncertainty assessment.

Further, we proposed a sparse implementation using a Least Angle Regression approach to handle high-dimensional problems, which improves computational performance in dealing with high-dimensional stochastic problems. 
The proposed method demonstrates promising results for a wide range of problems with superior performance in numerical accuracy and computational efficiency. However, the method is only suited for sufficiently smooth responses. Extension to non-smooth problems is perhaps possible with local \PCE{} \cite{novak2023active}, which is a subject for future research. 
Also, the extension of the proposed method to handle larger, more practical problems in scientific machine learning and uncertainty quantification is an exciting area for further investigation.

\section{Acknowledgements}
This work was supported by the Defense Threat Reduction Agency, Award HDTRA12020001. LN gratefully acknowledges the support of the Czech Science Foundation under project No. 23-04974S. The international collaboration was supported by the Ministry of Education, Youth and Sports of the Czech Republic under project No. LUAUS24260.

\appendix

\section{Reduced PCE Illustration}
\label{sec:Reduced_PCE}

Consider the input vector containing a single spatial coordinate, a time coordinate, and a single random variable $\boldsymbol{X} = \left[x, t,\xi\right]^{\top}$ and applying second-order polynomial basis functions. Here, the index set is given by:
\begin{equation*}
    \pazocal{A}_{\boldsymbol{X}} = \{(0,0,0), (0,0,1), (0,1,0), (1,0,0), (0,1,1), (0,0,2), (1,1,0), (0,2,0), (1,0,1), (2,0,0) \}
\end{equation*}
Partitioning the elements as $\boldsymbol{\alpha}=\left(\boldsymbol{\alpha_{\pmb{\pazocal{X}}}}, \boldsymbol{\alpha}_{\boldsymbol{\xi}}\right)$, we obtain the sets:
\begin{equation*}
    \pazocal{A}_{\boldsymbol{\xi}} = \{(0), (1), (2)\} \\
\end{equation*}
corresponding to the possible degrees of the third ($\xi$) variable, 
For each element of $\pazocal{A}_{\boldsymbol{\xi}}$, we can define the conditional sets as 
$\pazocal{T}_0 = \{(0,0), (0,1), (1,0), (1,1), (0,2), (2,0)\}$, $\pazocal{T}_1 = \{(0,0), (0,1), (1,0)\}$, and $\pazocal{T}_2 = \{(0,0)\}$.

Using the above-defined conditional sets, we can express $\pazocal{A}_{\boldsymbol{X}}$ as:
\begin{align*}
    \pazocal{A}_{\boldsymbol{X}} &= \{\pazocal{T}_{\boldsymbol{\alpha}_{\boldsymbol{\xi}}} \times \pazocal{A}_{\boldsymbol{\xi}} \}\\
    &= \{ \pazocal{T}_0 \times (0), \pazocal{T}_1 \times (1), \pazocal{T}_2 \times (2) \}
\end{align*}

Using Eqs. \eqref{eq:redPC^2-2} and \eqref{eq:redPC^2-3}, the reduced \PC{} expansion is given as 
\begin{equation*}
    Y_{\mathrm{PC}^2}(\xi | \ x,\ t) =  {y}_{(0)} \left(x,\ t \right){\Psi}_{(0)}(\xi) + {y}_{(1)} \left(x,\ t \right){\Psi}_{(1)}(\xi) + {y}_{(2)} \left(x,\ t \right){\Psi}_{(2)}(\xi),
\end{equation*}
where 
\begin{align*}
    {y}_{(0)} (x,\ t) =& {y}_{(0,0,0)}{\Psi}_{(0,0)}(x,\ t) +
    {y}_{(0,1,0)}{\Psi}_{(0,1)}(x,\ t)+
    {y}_{(1,0,0)}{\Psi}_{(1,0)}(x,\ t)+\\
    &{y}_{(1,1,0)}{\Psi}_{(1,1)}(x,\ t)+
    {y}_{(0,2,0)} {\Psi}_{(0,2)}(x,\ t)+
    {y}_{(2,0,0)} {\Psi}_{(2,0)}(x,\ t),\\
    {y}_{(1)} (x,\ t) =& {y}_{(0,0,1)}{\Psi}_{(0,0)}(x,\ t) +
    {y}_{(0,1,1)} {\Psi}_{(0,1)}(x,\ t)+
    {y}_{(1,0,1)} {\Psi}_{(1,0)}(x,\ t),\\
    {y}_{(2)} (x,\ t) =& {y}_{(0,0,2)}{\Psi}_{(0,0)}(x,\ t).
\end{align*}
\bibliographystyle{ieeetr}
\bibliography{references}  






\end{document}